\documentclass[journal]{IEEEtran}

\usepackage{subcaption}
\usepackage{multirow}
\usepackage{balance}
\usepackage{hyperref}
\usepackage{dsfont}
\usepackage{breqn}
\usepackage{makecell}
\usepackage{xcolor}
\usepackage{amsfonts}

\title{ARCSnake: Reconfigurable Snake-Like Robot with Archimedean Screw Propulsion for Multi-Domain Mobility}
\author{Florian Richter$^{\dagger,1}$ \IEEEmembership{Student Member, IEEE} , Peter V. Gavrilov$^{\dagger,2}$, Hoi Man Lam$^{\dagger,2}$,\\ Amir Degani$^3$ \IEEEmembership{Member, IEEE}, and Michael C. Yip$^1$ \IEEEmembership{Senior Member, IEEE}
\thanks{$^\dagger$ Equal Contribution}
\thanks{$^1$Florian Richter and Michael C. Yip are with the Department of Electrical and Computer Engineering, University of California San Diego, La Jolla, CA 92093 USA. {\tt\small \{frichter, yip\}@ucsd.edu}}
\thanks{$^2$Peter V. Gavrilov and Hoi Man Lam are with the Department of Mechanical and Aerospace Engineering, University of California San Diego, La Jolla, CA 92093 USA. {\tt\small \{pgavrilo, hml024\}@ucsd.edu}}
\thanks{$^3$A. Degani is with the Faculty of Civil and Environmental Engineering and with the Technion Autonomous Systems Program, Technion – Israel Institute of Technology, Haifa 3200003 Isreal. {\tt\small adegani@technion.ac.il }}
}

\begin{document}

\maketitle

\begin{abstract}
Exploring and navigating in extreme environments, such as caves, oceans, and planetary bodies, are often too hazardous for humans, and as such, robots are possible surrogates.
These robots are met with significant locomotion challenges that require traversing a wide range of surface roughnesses and topologies.
Previous locomotion strategies, involving wheels or ambulatory motion, such as snake platforms, have success on specific surfaces but fail in others which could be detrimental in exploration and navigation missions.
In this paper, we present a novel approach that combines snake-like robots with an Archimedean screw locomotion mechanism to provide multiple, effective mobility strategies in a large range of environments, including those that are difficult to traverse for wheeled and ambulatory robots.
This work develops a robotic system called ARCSnake to demonstrate this locomotion principle and tested it in a variety of different terrains and environments in order to prove its controllable, multi-domain, navigation capabilities.
These tests show a wide breadth of scenarios that ARCSnake can handle, hence demonstrating its ability to traverse through extreme terrains.
\end{abstract}


\section{Introduction}
Search and rescue missions, as well as research and exploration, can comprise of extreme environments that are too treacherous for humans to safely or easily traverse. 
One such example is the 2018 Tham Luang cave rescue where a soccer team was entrapped in a cave due to flooding \cite{bbc_2018}. 
Rescuers navigated through dangerously narrow caves, often diving through flooded portions that are too tight for the divers to wear their own scuba tanks.
Another example is NASA's proposed mission to search for extant life in the subterranean ocean of Saturn's Moon, Enceladus \cite{spencer_niebur_2010}.
A probe must travel across the harsh icy surface and then grip and power its way down tortuous, active plume vents to reach the subterranean ocean below.
These missions provide an inherently hostile environment for humans, and as such require a versatile robotic platform to challenge the multitude of terrains and environments, including both aquatic and terrestrial.
To achieve this, the robot must offer multiple consistent locomotion options to adapt to the terrain, flexibility to conform to the environment, and a small cross-section to access tight spaces.

\begin{figure}[t!]
    \vspace{2mm}
    \centering
    \includegraphics[width=0.96\columnwidth, trim={0 0 0 0 mm}, clip=true]{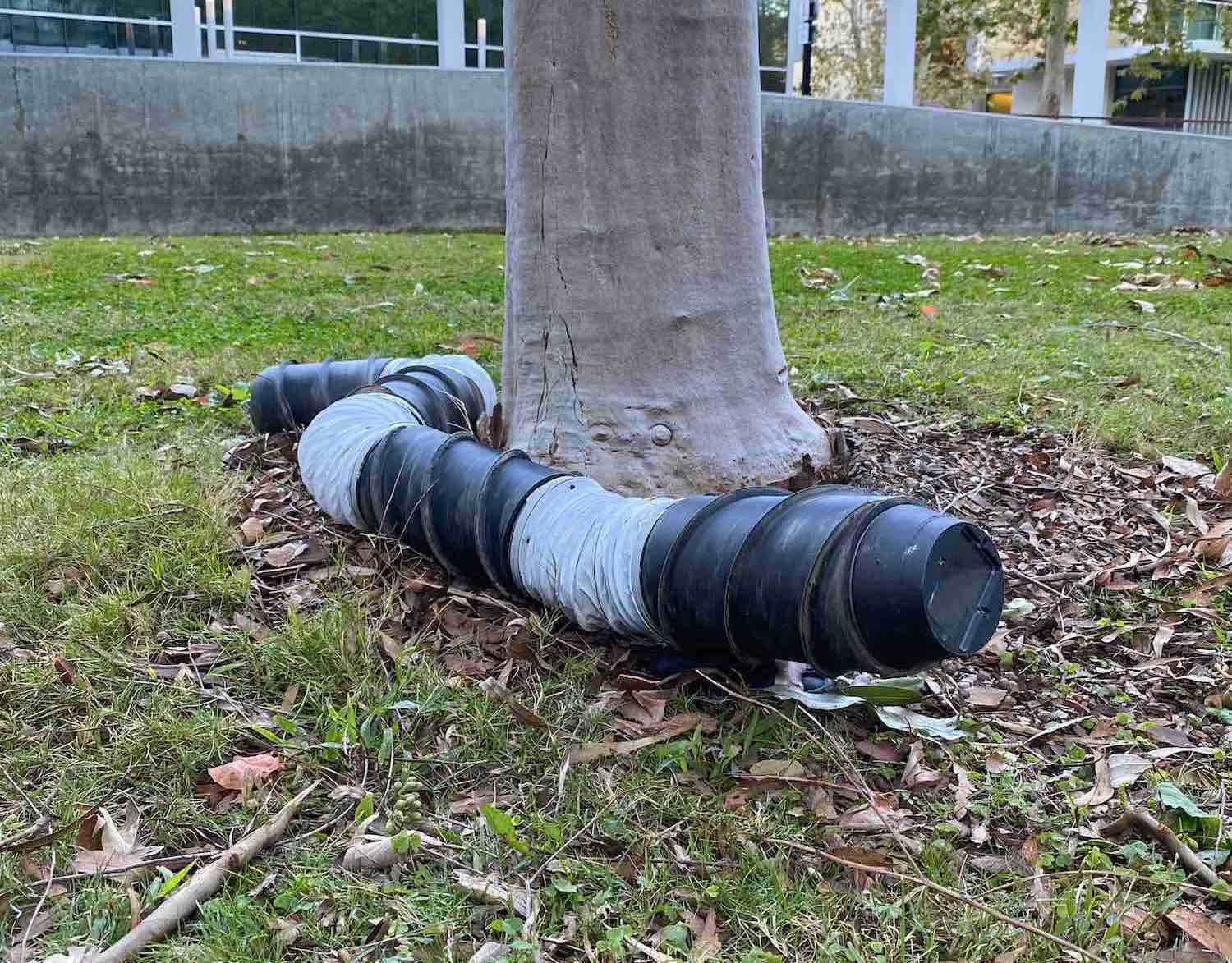}
    \caption{ARCSnake is a reconfigurable snake-like robot with Archimedean screws to provide propulsion on a wide range of terrain. In addition, its kinematic hyper-redundancy provides the wide library of bio-inspired locomotion methods from snakes. This culmination of unique control strategies in ARCSnake make it a robotic platform that meets the needs of exploration and search and rescue missions.}
    \label{fig:cover}
\end{figure}

In nature, snakes uniquely have the ability to conform to a variety of environments and navigate through rough terrains while maintaining a small cross-section.
They achieve this by utilizing an array of locomotion methods: undulating like a wave to propel over uneven ground, sidewinding to shift their body over dunes, and rectilinear tunneling through constricted spaces by shifting their skin \cite{gray1946mechanism,jayne1986kinematics,Newmanjeb166199}.
To mimic these capabilities, robotic designs were developed with hyper-redundant kinematic chains \cite{hirose_1992,hirose2004biologically, shammas2006three, wright2007design, zhang2020environmental}, allowing for traversal through tortuous narrow passageways such as pipes \cite{trebuvna2016inspection}.
This unique ability has made them an anticipated platform to use for search and rescue \cite{transeth2009survey,wright2012design}.

\subsection{Related Works}

Robotic replication of the undulations found in snakes proved successful in aquatic environments \cite{crespi2006amphibot,kamamichi2006snake}.
However, for land-based terrains, the challenge arises from producing enough force to propel the body forward due to the limited points of contact with the surface \cite{shugen2001analysis,ma2006analysis,bayraktaroglu2006design,marvi2014sidewinding} resulting in inconsistent propulsion when on low friction or granular terrain.
For more consistent surface contact or propulsion, snake robot designs with \textit{passive skins}, such as wheels and fins, have been explored.
One popular example is passive rollers which allow for consistent contact on smooth surfaces \cite{ye2004turning,crespi2005amphibot,wu2010cpg, cao2017robust}.

An under-explored area for robotic snake propulsion is the use of \textit{active skins} which produce unique non-snake-like locomotions.
Active skins provide propulsion for the snake while the hyper-redundant kinematic chain will be used to ensure consistent contact with the surface and control the direction of resulting locomotion. An initial approach was demonstrated in \cite{klaassen1999gmd,date2000locomotion} that used active wheels.
This approach required the robot to manage both the kinematic complexity from the snake-like structure while controlling the direction of the wheel propulsion force.
To bring about omnidirectional driving, omni-directional wheels as active skins have been explored \cite{ye2009modular}, and Fukushima et al. developed passive rollers on actively rotating bodies, similar to Mecanum wheels  \cite{fukushima2012modeling}.
These designs however are limited to smooth, planar surfaces, hence not feasible for exploration nor search and rescue missions.
For aerial maneuvers, articulated rotors have been designed to allow for controlled motions in air \cite{zhao2018design,zhao2018flight}.
For propulsion on rigid terrains with non-smooth surfaces, toroidal skins \cite{mckenna2008toroidal} and tank treads \cite{armada2005omnitread,borenstein2007omnitread} were previously presented. 
While these skins are effective in their respective specialized terrains, they fail to generalize to others.

In this work, we present a novel type of active skin for snake robots, the Archimedean screw, which allows for consistent propulsion on a wide breadth of terrains. 
While the concept of the Archimedean screw is not new (in fact, originating over 2000 years ago), these screws have only recently been considered for mobility. In the robotics space, groups have shown Archimedean screw propulsion for mobile robots / vehicles as a method to traverse over rough or viscous terrain \cite{nagaoka2009development,osinski2015small,lugo2017design,he2017design}. The Archimedean screw provided consistent propulsion across a variety of extreme terrain including water, marshes, soft soil, snow, and ice \cite{neumeyer1965marsh,dugoff1967model,fales1972riverine}.
Although the Archimedean screw drive was capable of traversing terrain where other ``robust'' methods (such as wheels and tank treads) failed, screw-driven platforms never became popularized due to their limited or lack of mobility on smooth rigid surfaces, such as on concrete.
This occurs when the surface has no material for the screws to propel through, which can be thought of as screw-slippage \cite{nagaoka2010terramechanics}.
Quad-screw designs have been proposed to take advantage of the partial screw-slippage case \cite{freeberg2010study,lugo2017conceptual}, but these designs are unable to conform to the environment for consistent contact on non-planar surfaces.
Furthermore, the conditions for partial screw-slippage are very specific, and instead, our proposed mobility strategies rely on either screw propulsion or screw-slippage resulting in consistent locomotion.

\subsection{Contributions}

Our proposed mobile platform, ARCSnake, combines the Archimedean screw with the flexibility of snake-like robots to retain the advantages of screw propulsion and overcome the slippage challenges by using its hyper-redundancy.
We previously presented the embedded components of the ARCSnake \cite{schreiber2020arcsnake}; this work extends off that proof-of-concept and investigates the entire system from kinematic modeling to control, and presents extensive testing and analysis on multi-domain mobility.
To this end, we intend to make the following novel
contributions:
\begin{enumerate}
    \item A scalable mechanical, electrical, and software design for a screw-propelled, snake-like robot,
    \item Kinematic models and controllers for both Archimedean screw propulsion and screw-slippage cases resulting in a variety of locomotion schema, derived from active configurations of the snake body, and
    \item A teleoperation system for hyper-redundant robots to produce complex snake-inspired maneuvers.
\end{enumerate}
The large variety of experiments conducted with ARCSnake such as traversal through outdoor environments and obstacle-like challenges show that the locomotion strategies resulted by combining screw propulsion with snake-like, hyper-redundancy meets the needs of search and rescue, and exploration in extreme terrains.

{
\renewcommand{\arraystretch}{1.0}
\begin{table}[t]
    \vspace{2mm}
    \centering
    \caption{ARCSnake design specifications highlighting the redundant sensing and kinematic chain. A Body contains an inner shell and an Archimedes' screw.}
    \begin{tabular}{c|l|l}
    \textbf{Specifications} & \textbf{Within} & \textbf{Values} \\ \hline
    \multirow{2}{*}{Power} & Module & 12-60V, 310W (max)   \\ 
     &System & 12-60V, 1240W (max)\\ \hline
    \multirow{2}{*}{Communication} & Module & I2C \\
    &System & TCP/IP via ROS \\ \hline
    \multirow{6}{*}{Sensors} &  \multirow{4}{*}{Body} & Optical Encoder\\
                          && Motor Current\\
                          && 9 DoF IMU\\
                          && Temperature Sensor\\
     & \multirow{2}{*}{U-Joint} & Magnetic and Optical Encoders\\
     && Motor Current\\ \hline
     \multirow{2}{*}{Actuators} & Screw & Torque: 1.6Nm continuous, 2.0Nm peak\\
     & U-Joint & Torque: 2.1Nm continuous, 2.7Nm peak\\ \hline
      \multirow{3}{*}{Dimensions} & Body & Max Len: 19.6cm \  Max Dia: 12.5cm\\
      & U-Joint & Max Len: 16.8cm \  Max Dia: 11.0cm\\
     & System & Max Len: 128.7cm Max Dia: 12.5cm\\   \hline
     \multirow{3}{*}{Mass} & Module & Body: 1.0kg, U-Joint: 0.88kg\\ 
     & Head & 0.68kg \\
     & System & 6.1kg \\
    \end{tabular}
    \label{tab:specs}
\end{table}
}

\section{System Design}

The design of ARCSnake is split into three components: mechanical, electrical, and software.
Each component is designed with scalability in mind such that system can be extended to an arbitrary number of chained segments.
The manufactured and tested design is 4 segments long, and Table \ref{tab:specs} shows some of the key specifications of it.
However, it is challenging to interface and develop control strategies with robotic platforms that have many degrees of freedom.
Therefore, we also present a novel teleoperation system for hyper-redundant robots which brings about intuitive operator control.
This section will describe the (A) mechanical design, (B) electrical design, (C) software architecture, and (D) teleoperation scheme for the system.
The next sections will discuss kinematic modeling and control under different proposed locomotion schema.

\subsection{Mechanical Design}

\begin{figure}[t!]
    \vspace{2mm}
    \centering
    \includegraphics[width=0.9\columnwidth, trim={0 0 0 0 mm}, clip=true]{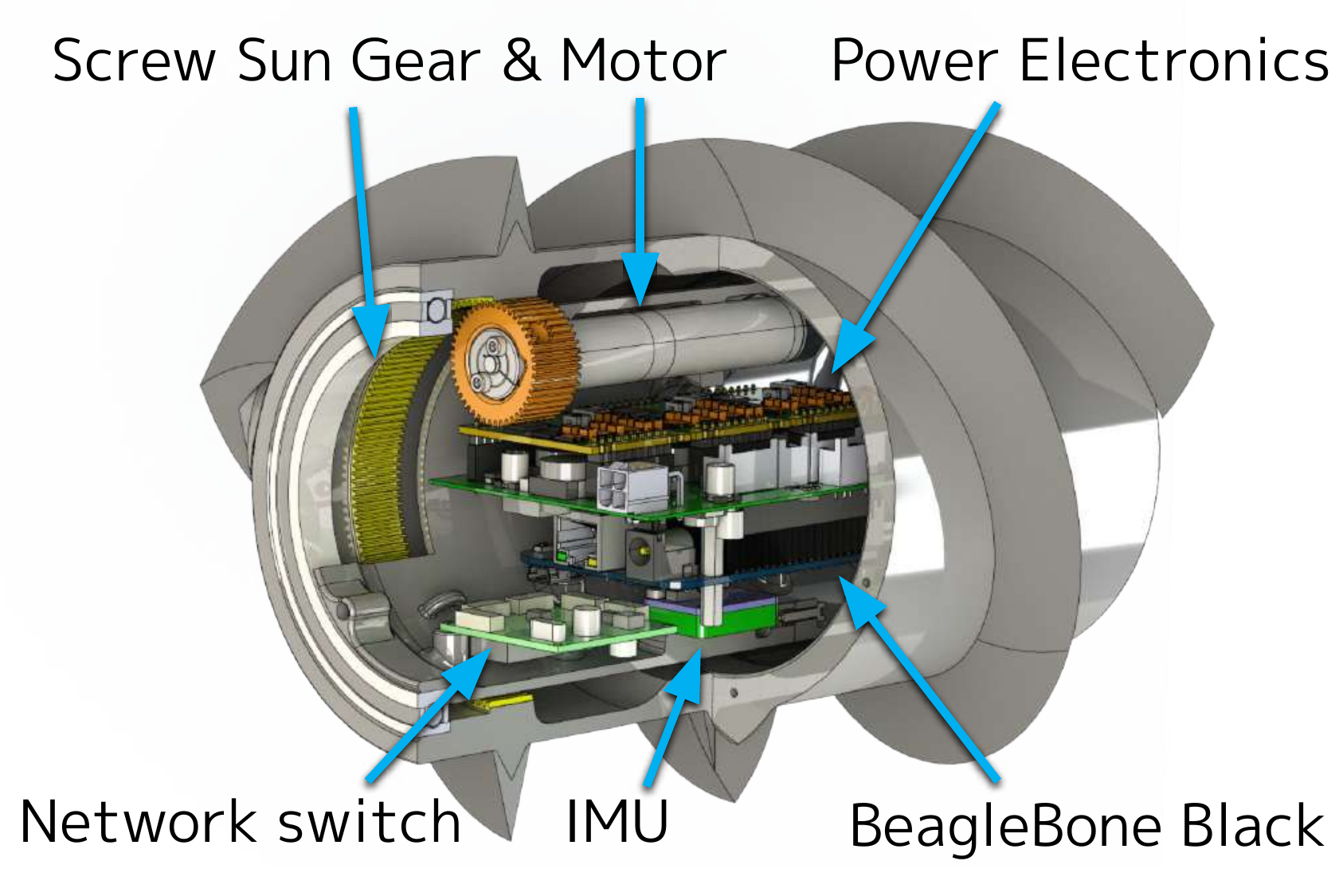}
    \caption{Cutout view of the Archimedean Screw segment. Each segment is self-contained by holding its own power electronics and embedded system for motor position and velocity regulation in the inner shell. The inner shell drives the outer Archimedean screw with a motor and sun gear.}
    \label{fig:segment}
\end{figure}

\begin{figure}[t!]
    \vspace{2mm}
    \centering
    \includegraphics[width=0.98\columnwidth, trim={10 0 10 0 mm}, clip=true]{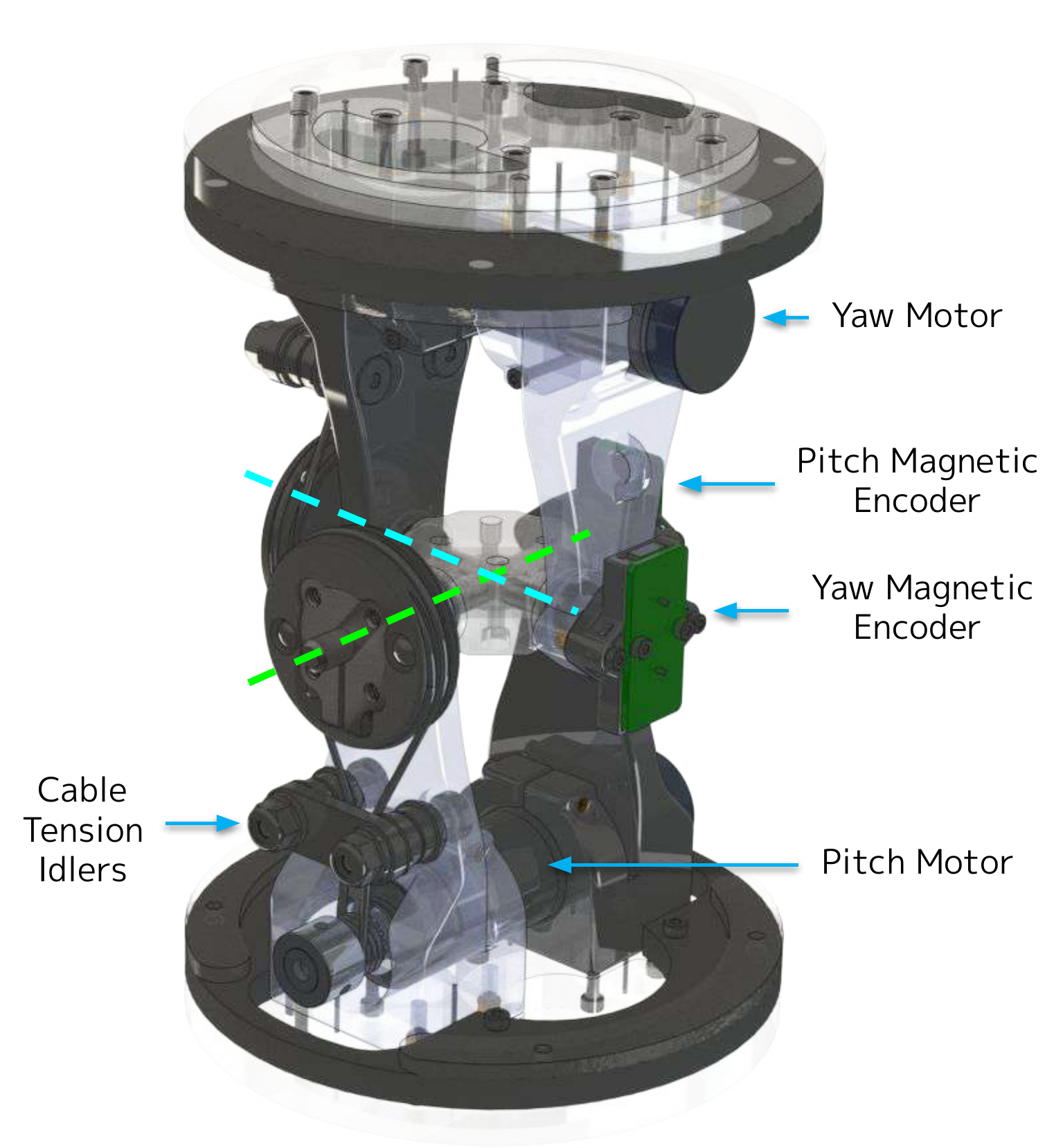}
    \caption{Rendering of a cable-driven U-Joint which connects the screw segments together, hence providing the hyper-redundancy for the system in a snake-like manner. Magnetic encoders are used on the U-Joint to provide absolute positioning.}
    \label{fig:U_Joint}
\end{figure}

Each ARCSnake segment is composed of an inner core housing power electronics and embedded systems, surrounded by an Archimedean screw as shown Fig. \ref{fig:segment}. Each screw segment is serially connected by universal joints (U-Joint) which is shown in Fig. \ref{fig:U_Joint}. The
ARCSnake prototype is composed of four repeated units, with the foremost unit's joint replaced by a camera.
In practice, the snake body can get longer with more active segments due to its serially chainable and modular design but generally cannot get shorter than four segments without loss of locomotion capability.

The Archimedean screw shell is designed with screw threads that have a 22$^\circ$ pitch angle and a two-screw start to increase surface contact.
This pitch angle has been experimentally shown to maximize drawbar pull and minimize slippage for mobile locomotion with Archimedean screws \cite{archimedes_ref}.
A smaller default screw thread height is selected for the screw shell with a diameter of 128mm. 
The screw is driven using a sun gear and an ECXSP16L motor with GPX16HP 35:1 gearhead.
The result is a screw lead speed of 0.23m/s with 1.6Nm and 2.0Nm continuous and peak torque, more than enough torque to pull its own body weight over challenging terrain.

The U-Joint has a hemispherical range of motion with just over 180$^\circ$ of range in both pitch and yaw axis.
Each axis is cable-driven with a 3.8:1 reduction by an ECXSP16L motor with a GPX16HP 35:1 gearhead, resulting in a measured continuous torque output of 3.4 Nm, which is enough to lift its own body weight (i.e., an adjacent link).
AS5048B I2C 14-bit Magnetic encoders are mounted on both pitch and yaw axes, which are used in combination with the motor's optical encoders for redundant sensing.
The U-joint is designed for control in the pitch-yaw axis, rather than a pitch-roll axis because a pitch-roll axis is singular when the snake is straight, which is a frequent configuration for conforming to narrow spaces.
Instead, for the pitch-yaw configuration, the robot has the highest controllability in the straight configuration, and only loses controllability when the robot is fully curled up (i.e. reaches 90$^\circ$ angles on the U-Joints).
This presents a much more friendly kinematic layout for snake-like maneuvers.

\begin{figure*}
    \centering
    \includegraphics[width=0.98\textwidth]{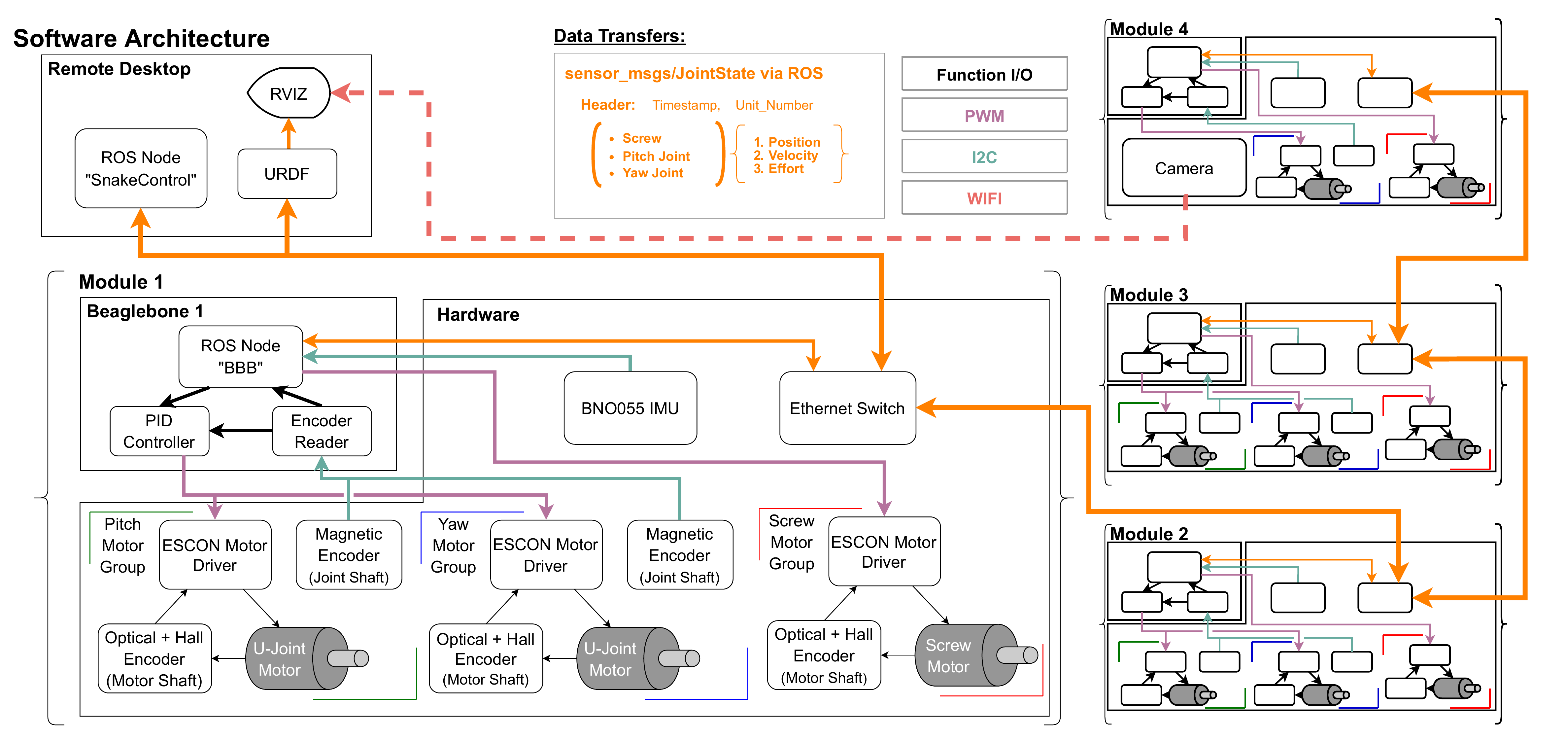}
    \caption{Complete software architecture implemented on ARCSnake. The BeagleBone Blacks regulate motors locally on the segment while broadcasting sensor information over network. Meanwhile, a remote desktop receives the data and sends target set-point commands back to the BeagleBone Blacks. }
    \label{fig:software_architecture}
\end{figure*}

\subsection{Electrical Design}

Each segment is self-contained by holding its own set of power electronics, sensors, and an embedded system for local control and communication.
BeagleBone Black is chosen for the embedded system \cite{beagle_bone}, and a custom-printed circuit board (PCB) is designed to hold all the necessary power electronics.
The power electronics include three high-efficiency buck-boost regulators, VICOR PI3740-00, for a 24V internal rail and a step-down regulator Texas Instrument's TPS54561, followed by a low dropout 3.3V regulator, Texas Instruments TL5209.
The resulting regulated 3.3V powers the BeagleBone Black.
The internal 24V rail also provides power to three onboard motor controllers, Maxon Motor ESCON 50/5, which drive the segments' respective U-Joint and screw motors.
By using the buck-boost regulators, all the subsequent electronics are robust against any voltage sags which could occur from long cable runs.
Therefore, a power line is daisy-chained along ARCSnake with no losses in electronic capabilities.
The power line is set to 24V as that is where VICOR PI3740-00 is most efficient.

All segments additionally house a network switch to daisy-chain an Ethernet cable for communication between other segments and a master PC.
On a vibrational dampening rubber, a BNO055 Inertial Measurement Unit (IMU) is also housed in each segment to provide a world reference orientation.
These ancillary components and magnetic encoders from the corresponding U-Joint are powered off the low dropout 3.3V regulator from the custom PCB.
This self-contained electronic design allows for any number of segments to be chained together with ease.

\subsection{Software Architecture}
The software architecture, similar to the electrical design, is meant for scalability such that an arbitrary number of segments can be supported.
To this end, each BeagleBone Black regulates the position of the U-Joint motors using the magnetic encoders for position feedback.
Meanwhile, the screw motor's velocity is regulated by the ESCON Motor driver, and the BeagleBone Black simply sends set-points to the driver.
A remote desktop, which would run a higher-level controller, sends set-point positions and velocities to the BeagleBone Blacks via the Robot Operating System (ROS) \cite{ros}.
All sensor data is also sent over ROS by the BeagleBone Blacks to the remote desktop.
Lastly, a camera that is attached to the head of the snake feeds visual data over Wi-Fi to the remote desktop.
A flowchart of this scalable architecture is shown in Fig. \ref{fig:software_architecture}.

\subsection{Teleoperation using Voodoo Doll}

\begin{figure}[t]
    \centering
    \includegraphics[width=0.48\textwidth]{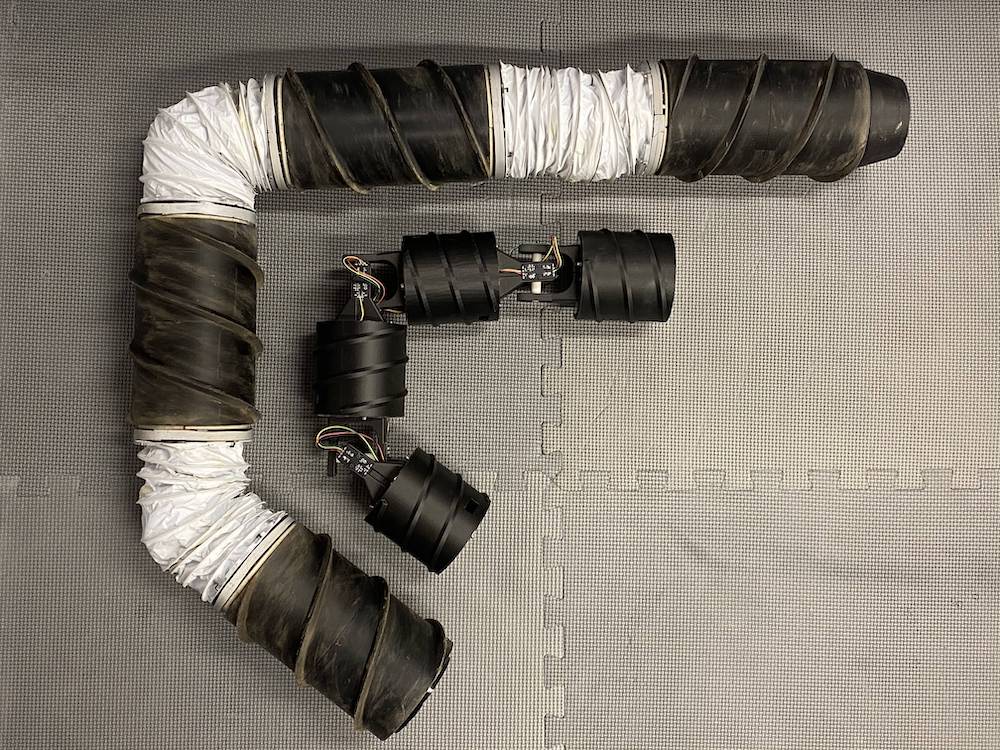}
    \caption{Voodoo Doll is a scaled model of ARCSnake and used as a control input device for an operator of ARCSnake. Use of Voodoo Doll allows the operator to react quickly to unexpected disturbances, plan trajectories on the fly, and abstract away explicit U-Joint angles on ARCSnake.}
    \label{fig:voodoo_real_world}
\end{figure}

\begin{figure}[t]
    \centering
    \includegraphics[trim=20cm 2cm 5cm 0cm, clip, width=0.425\textwidth]{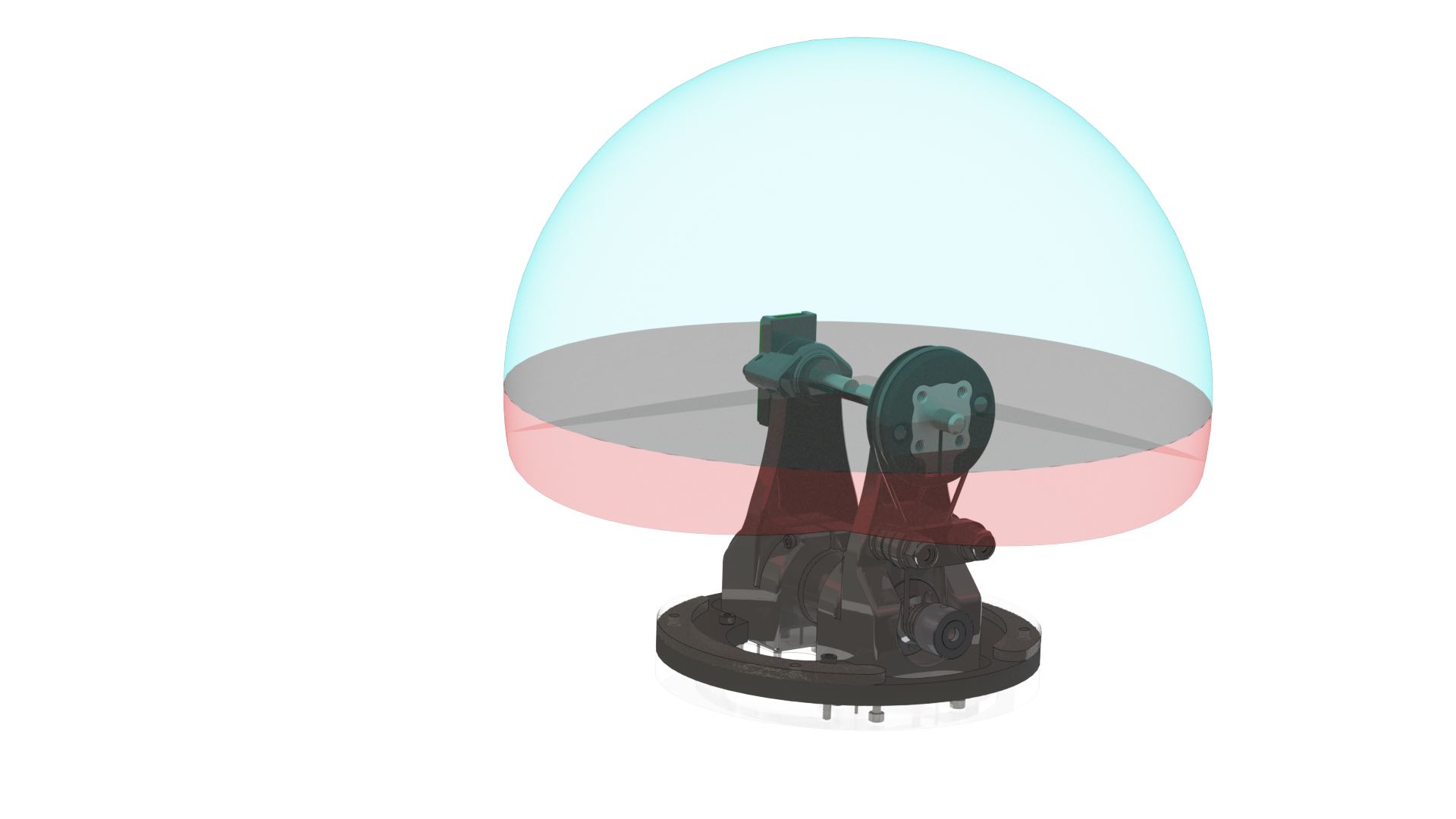}
    \caption{To prevent the operator from inputting extreme joint setpoints or crossing joint singularities, Voodoo Doll has mechanical limits to form a subset of ARCSnake’s joint configuration space. In this figure, the red dome represents ARCSnake’s distal joint movement range and the blue dome represents the Voodoo Doll’s reduced convex configuration space.}
    \label{fig:joint_limits_voodoo}
\end{figure}

To understand and test ARCSnakes abilities, a specialized teleoperation device was developed for intuitive control of the complex, hyper-redundant system.
We coin this device \textit{Voodoo Doll} because it is a scaled-down model of the actual ARCSnake and gives a one-to-one kinematic mapping between the operator’s device and ARCSnake’s U-Joints, as shown in Fig. \ref{fig:voodoo_real_world}.
Each U-Joint on the Voodoo Doll has two magnetic encoders (AS5048B) to measure the operator's command inputs.
The magnetic encoders are daisy-chained via I2C and a 3.3V power line.
The values are continuously read by a BeagleBone Black and sent as set-points to the corresponding U-Joints on ARCSnake with negligible delay.
This allows for a user to puppet the robot’s configuration remotely in a natural manner.
Furthermore, the low latency between the Voodoo Doll and ARCSnake allows for adaptation to changing terrains.
The joint limits on Voodoo Doll are slightly more restricted than on ARCSnake, as shown in Fig. \ref{fig:joint_limits_voodoo}, to prevent the user from crossing joint singularities and sending commands that exceed joint limits.

Voodoo Doll is particularly helpful in situations where adaptation to sudden environment changes is necessary.
One such use case is climbing ledges where the height and depth are unknown before testing.
The use of Voodoo Doll abstracts joint control away from specific U-Joint angles, allowing the operator to intuitively control joint movements in 3-D while focusing on environment conformation.
Furthermore, it allows the user to adapt and correct for disturbances in real-time.
This is in contrast to previous experiments where specialized testbeds were designed in order to highlight a snake robot's capabilities \cite{xiao2018review}.

\begin{figure*}[t]
    \centering
    \includegraphics[width=0.45\linewidth, trim=4.5cm  8.3cm 6.2cm 0, clip]{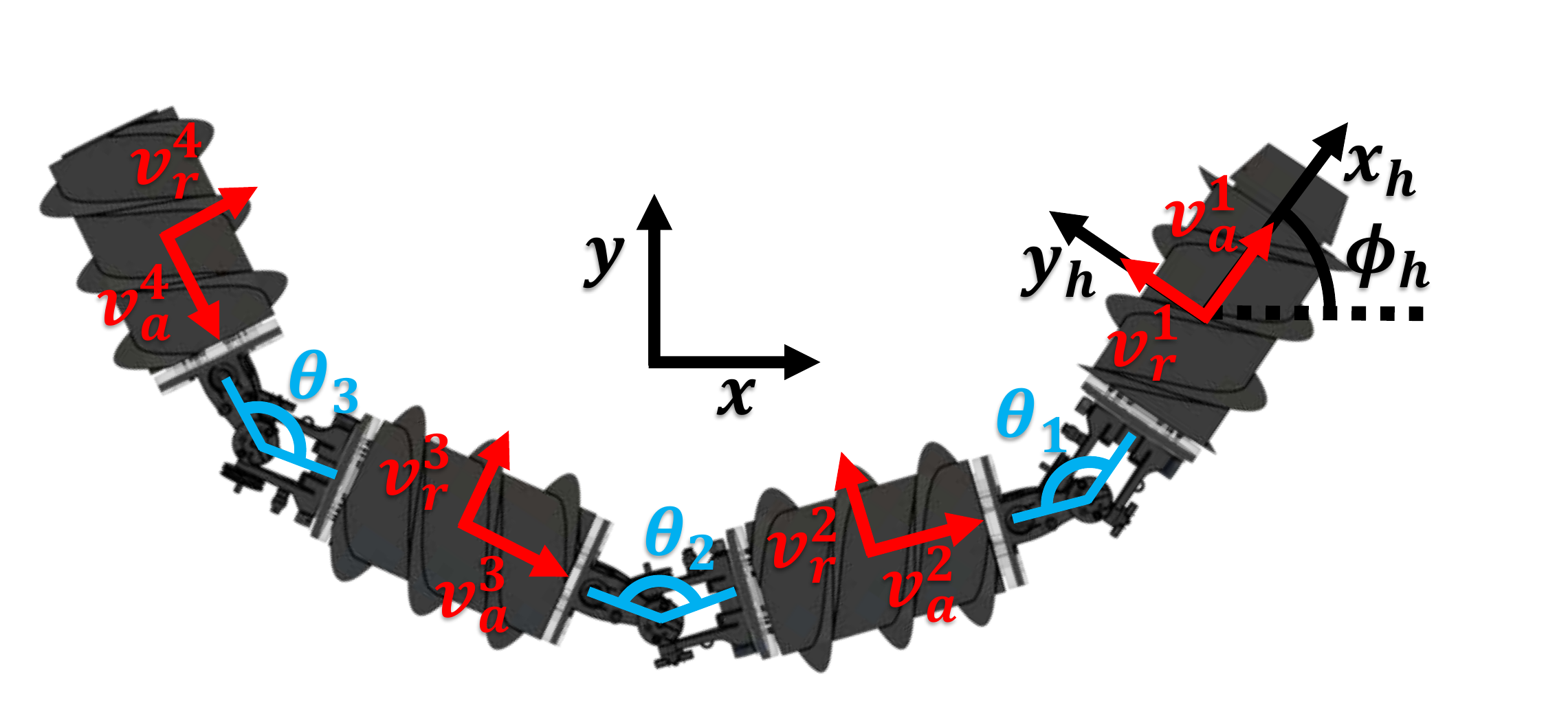}    
    \includegraphics[width=0.475\linewidth, trim=4.5cm  8.3cm 6.2cm 0, clip]{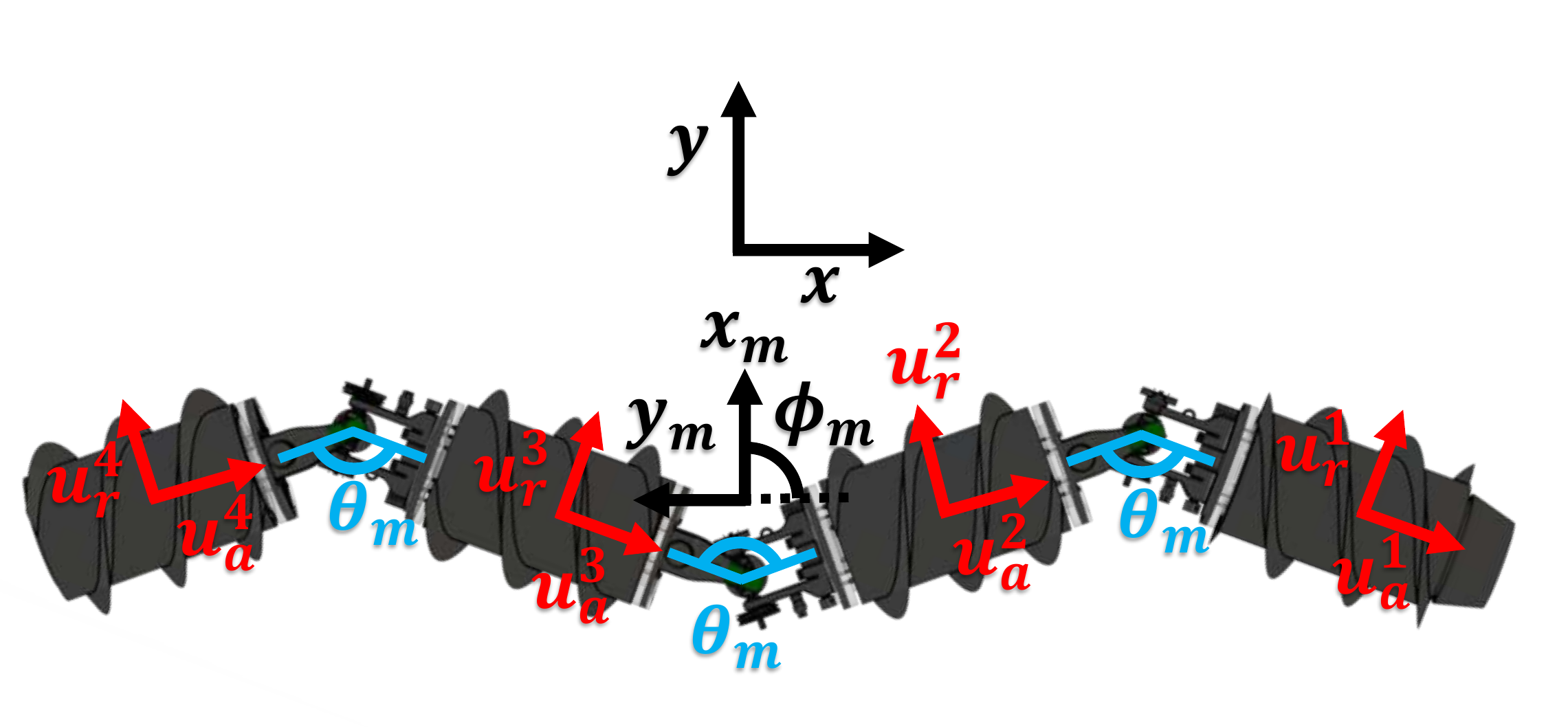}
    \caption{From left to right, visualization of the variables used for tunneling mode and M-configuration kinematic modelling. Red arrows highlight velocities at each segment, blue the planar joint angles, and black coordinate frames and heading.}
    \label{fig:kinematic_model}
\end{figure*}

\section{Kinematic Modelling}

While the Voodoo Doll will produce good control of the hyper-redundant kinematic chain, combining Archimedean screw propulsion to the different robot configurations requires kinematic models to determine efficient orientations and velocities for each Archimedean screw segment.
This differs from previous work on controlling robotic snakes such as gaits \cite{tesch2009parameterized, melo2014experimental, cao2020modeling} and central pattern generators \cite{bing2017towards} as they only utilize the hyper-redundancy for locomotion which ARCSnake does have, and hence these methods can be applied to ARCSnake.
Two new locomotion modes based on the Archimedean screw propulsion are proposed with associated kinematic models: \textit{tunneling mode} and \textit{M-configuration}; they handle two separate cases, screw engagement and screw-slippage respectively.
In scenarios where the screws slip, the resulting motion of the Archimedean screw is that of a wheel.
Examples of such terrain are where previous screw-robots typically fail as there is no material to push on for propulsion.
Here, ARCSnake can reconfigure its body to an M-configuration to still achieve effective locomotion.

\subsection{Tunneling Mode}
In tunneling mode, ARCSnake relies on screw propulsion for traversal to minimize its overall cross-sectional area and can set the U-Joints to adjust heading. For the planar model of ARCSnake as shown in Fig. \ref{fig:kinematic_model}, the center location of segment $i$, $(x^i_h, y^i_h)$, in the head segment's coordinate frame (denoted with a subscript $h$), is:
\begin{equation}
    \begin{bmatrix} x^i_h \\ y^i_h \end{bmatrix} = 
    \begin{bmatrix} -l - \sum \limits_{j=1}^{i-1} \left( 2l \cos \left(\theta^\prime_{1:j} \right) \right)  + l \cos \left(\theta^\prime_{1:i-1} \right) \\ \sum \limits_{j=1}^{i-1} \left( 2l \sin \left( \theta^\prime_{1:j} \right) \right)  - l \sin \left( \theta^\prime_{1:i-1} \right) \end{bmatrix}
\end{equation}
where $l$ is the distance from $(x^i_h, y^i_h)$ to its neighboring U-Joint,
\begin{equation}
    \theta^\prime_{1:j} = \sum \limits_{k=1}^j (\pi - \theta_{k})
    \label{eq:summation_short_notation}
\end{equation}
and $\theta_{k}$ is the $k$-th planar U-Joint angle.
Using this relationship, the velocity at the center location of segment $i$, $w^i_h \in \mathbb{R}^2$, induced by the U-Joints in the head segment's coordinate frame can be computed as:
\begin{equation}
    w^i_h =
    \begin{bmatrix}
        \sum \limits_{j=1}^{i-1} \left( 2l \sin \left(\theta^\prime_{1:j} \right) \dot{\theta}_{1:j} \right)   - l \sin \left(\theta^\prime_{1:i-1} \right)  \dot{\theta}_{1:i-1} \\
        \sum \limits_{j=1}^{i-1} \left( 2l \cos \left(\theta^\prime_{1:j} \right) \dot{\theta}_{1:j} \right)  - l \cos \left(\theta^\prime_{1:i-1} \right) \dot{\theta}_{1:i-1}
    \end{bmatrix}
\end{equation}
where
\begin{equation}
    \dot{\theta}_{1:j} = \sum \limits_{k=1}^j \dot{\theta}_k
\end{equation}
and $\dot{\theta}_k$ is the $k$-th planar U-Joint velocity.
Let ARCSnake's instantaneous velocity and change in heading be $\begin{bmatrix} \dot{x}_h & \dot{y}_h \end{bmatrix}^\top$ and $\dot{\phi}_h$ respectively in the head segment's frame.
Therefore, the angular velocity of the snake is $\omega = \dot{\phi}_h$ and the center of rotation is $(0, \dot{x}_h/\dot{\phi}_h)$ in the head segment's frame.
The axial and radial velocity at segment $i$ is computed as:
\begin{equation}
    \begin{bmatrix}
        v_{a}^i \\ v_{r}^i
    \end{bmatrix} = \mathbf{R}\left( -\theta^\prime_{1:i-1} \right) \left(
    \begin{bmatrix}
     \dot{x}_h - y^i_h \dot{\phi}_h  \\\dot{y}_h + x^i_h  \dot{\phi}_h
    \end{bmatrix} + w^i_h\right)
\end{equation}
where
\begin{equation}
    \mathbf{R}(\theta) = \begin{bmatrix} \cos(\theta) & -\sin(\theta) \\
    \sin(\theta) & \cos(\theta) \end{bmatrix} 
\end{equation}
the standard 2D rotation matrix.
Hence, completing a general kinematic relationship for the system.

In tunneling mode, ARCSnake keeps the U-Joints at fixed locations and relies on screw propulsion for traversal to minimize its overall cross-sectional area.
Therefore, the constraint of $\dot{\theta} = 0$ and $v_r^i = 0$ is applied to the kinematic relationship in order describe the fact that only screw propulsion is used.
The first constraint implies that $w^i_h = 0$ and the joint angles are regulated to hold a constant position.
In the physical world, the second constraint is met on surfaces where no wheel-like propulsion occurs, such as aqueous environments.
On surfaces where wheel-like propulsion can occur, such as grass, ARCSnake will cancel them out using the alternating handedness of the screws.
This constraint allows for solutions of U-Joint angles and axial velocities given a target set of velocities, $\dot{x}_h, \dot{\phi}_h$, for the robot achieve.
Note that the constraint forces $\dot{y}_h = 0$.
In practice, however, some of these solutions are infeasible to reach due to limitations of the ability to control the axial velocity, $v^i_a$.
Due to the screw's high-reliance on surface contact and terrain, the velocity a segment reaches given a screw velocity is difficult to determine.
Therefore, the problem is instead simplified to solving for the set of joint angles that give a turning radius, $\dot{x}_h/\dot{\phi}_h$ and assume all screws are equally propelling, $v^i_a = v^j_a$.
Then the solution is that all the joint angles are set to the same value, $\theta$, such that the tangents to $v^i_a$ intersect at the center of rotation, $(0, \dot{x}_h/\dot{\phi}_h)$.
The angle is computed as:
\begin{equation}
    \theta = 2 \tan^{-1}\left( \frac{\dot{x}_h/\dot{\phi}_h}{l} \right)
\end{equation}
Regardless of difficulties in determining the magnitude of the resulting axial velocity, $v^i_a$, this model allows for controllability of the turning radius, so it can be used to construct paths with the kinematic constraint of $\dot{y}_h = 0$. 
While the self-imposed constraint of $v^i_r = 0$ limits the possible configurations for tunneling mode, it does provide a consistent method for control across all screw-propellable media.

\subsection{M-Configuration}


\begin{figure*}[t]
    \centering
    
    \includegraphics[width=0.6\columnwidth]{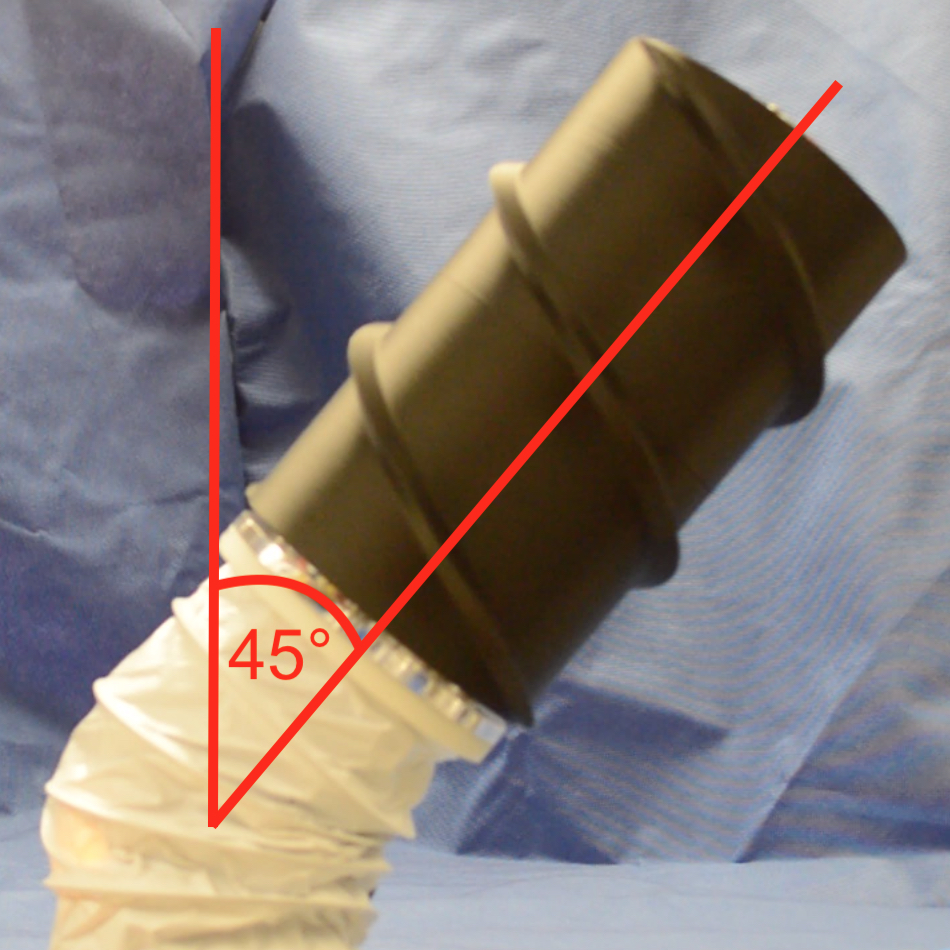}
    \hspace{1mm}
    \includegraphics[trim=0cm 1.25cm 0 0cm, clip, width=0.665\columnwidth]{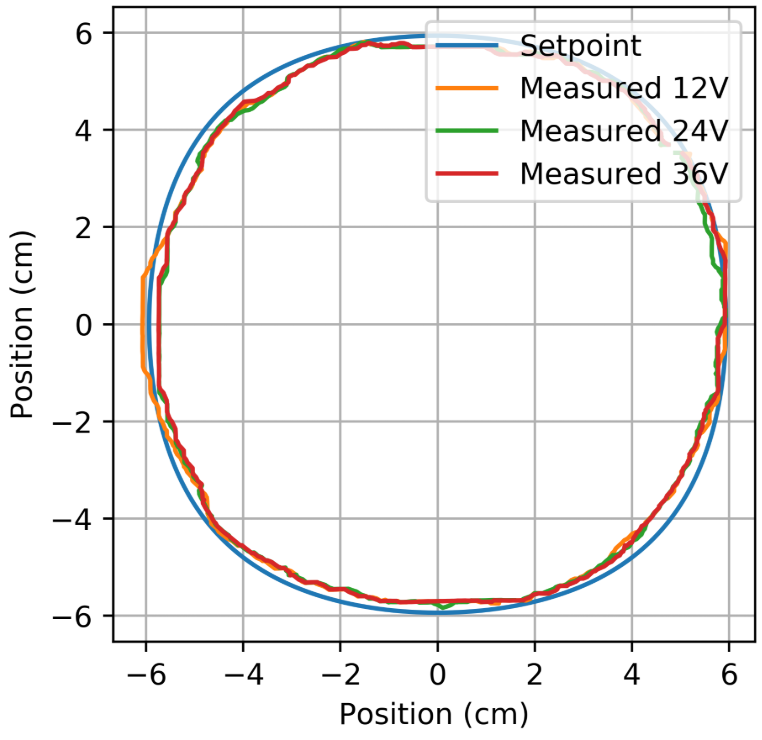}
    \hspace{4mm}
    \includegraphics[trim=0cm 1cm 0cm 1cm, clip, width=0.435\columnwidth]{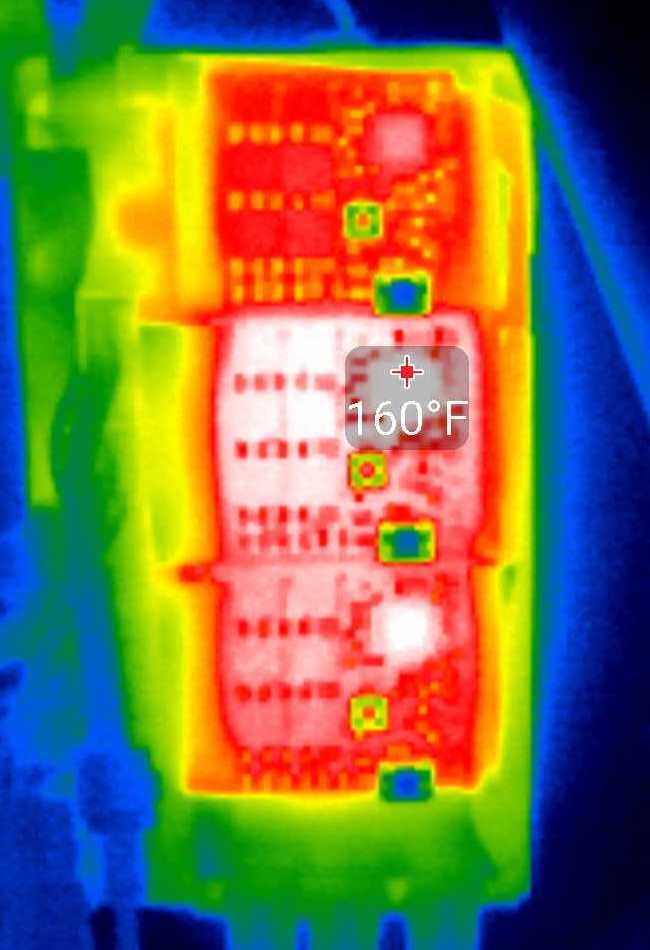}
    
    \caption{A module is placed in an inverted pendulum configuration shown in the left figure. The U-Joint regulates the body position in a circular trajectory across varying voltages which are displayed in the middle plot.
    The body position is measured using the magnetic encoders.
    As shown in the right, a thermal image is captured of the power electronics, and a maximum temperature of 71$^\circ$C is measured when the U-Joint is under heavy load.
    Both results highlight the good thermal performance and robustness to voltage disturbances of the electronic design.
    }
\label{fig:regulating_different_voltages}
\end{figure*}

In the M-configuration, a new coordinate frame (denoted with $m$) is placed along the center U-Joint and in line with all center locations of the segments as shown in Fig. \ref{fig:kinematic_model}.
This coordinate frame is defined because it is convenient to describe velocities with respect to the center of ARCSnake in the M-configuration.
The center location of segment $i$ in this new coordinate frame, denoted by $(x^i_m, y^i_m)$, is:
\begin{equation}
    \begin{bmatrix}
        x^i_m \\ y^i_m
    \end{bmatrix} = 
    \begin{bmatrix}
        0 \\
        (-2i + 5) l \cos(\theta^\prime_m/2)
    \end{bmatrix}
\end{equation}
where $\theta_m^\prime = \pi - \theta_m$ and $\theta_m$ is the planar angle of the U-Joints in the M-configuration.
Let the instantaneous velocity and change of heading be $\begin{bmatrix} \dot{x}_m & \dot{y}_m \end{bmatrix}^\top$ and $\dot{\phi}_m$ respectively in the new coordinate frame.
Similar to the computations from the previous section, the axial and radial velocity at segment $i$ is computed as:
\begin{equation}
    \label{eq:kin_model_m_configuration}
    \begin{bmatrix}
        u_{a}^i \\ u_{r}^i
    \end{bmatrix} = \mathbf{R}\left( (-1)^{i-1} \theta^\prime_m/2 - \pi/2 \right) \begin{bmatrix}
        \dot{x}_m - y^i_m \dot{\phi}_m \\ \dot{y}_m
    \end{bmatrix}
\end{equation}
resulting in a general kinematic relationship for M-configuration.

The M-configuration is designed to utilize wheel propulsion to locomote on rigid surfaces, such as concrete, where the screws can not engage.
Setting $\theta_m = \pi$ and $\dot{y}_m = 0$ will result in $u^i_a = 0$ from the kinematic relationship hence being conditions which ensure only wheel propulsion is utilized.
The issue with setting $\theta_m = \pi$ is there is no leverage for the wheels to propel with and the inner cores will spin while ARCSnake stays stationary.
Therefore, the condition is modified to $\theta_m < \pi$ and $\dot{y}_m = 0$ to minimize the axial velocities while still providing the leverage necessary. 
The condition $\theta_m < \pi$ directly contributes to slippage for the wheel propulsion, and the slippage ratio for segment $i$ is:
\begin{equation}
    s^i = 1 - \frac{u^i_r}{\omega^i r_s}
\end{equation}
where $\omega_i$ is controlled angular velocity of the screw and $r_s$ is the radius Archimedean screw.
Combining the slippage relationship with (\ref{eq:kin_model_m_configuration}), the following inverse kinematic model can be derived for segment $i$:
\begin{equation}
    \omega^i = \frac{\cos \left( \theta^\prime/2  \right)(y^i_m \dot{\phi}_m - \dot{x}_m)}{r_s - r_s s^i}
\end{equation}
which gives the control values necessary to reach a target $\dot{x}_m$ and $\dot{\phi}_m$.
However, it is not always feasible to measure the slippage ratio, so instead a ratio for wheel velocities, which are measurable, can be found:
\begin{equation}
    \frac{\omega^i}{\omega^j} = \frac{y^i_m - \dot{x}_m/\dot{\phi}_m}{y^j_m - \dot{x}_m/\dot{\phi}_m}
\end{equation}
which gives the relative control values necessary to reach a turning radius of $\dot{x}_m/\dot{\phi}_m$ assuming that the slippage ratios are equal for all segments.
In a similar manner as the result from the previous section, this model allows for controllability of the turning radius given the slippage ratio is greater than 0.
Therefore, it can be used to construct paths with the kinematic constraint of $\dot{y}_m = 0$.

\section{Experiments and Results}

A series of experiments are conducted to test ARCSnake's system design, multi-domain mobility, and obstacle clearance capabilities, all of which are critical aspects for exploration and search and rescue missions.
A large variety of terrains are experimented on to highlight the many surfaces tunneling mode and M-configurations cover and validate their respective kinematic models.
Meanwhile, the last series of experiments shows the platform's ability to utilize its hyper-redundancy for complex tasks such as stair climbing via snake-inspired maneuvers.
Together, these experiments show the large breadth of terrains and environments the platform is able to uniquely traverse.

\subsection{System Validation}

\begin{figure}[t]
    \centering
    \includegraphics[trim=1.2cm 3.5cm 20.5cm 4cm, clip, width=0.39\columnwidth]{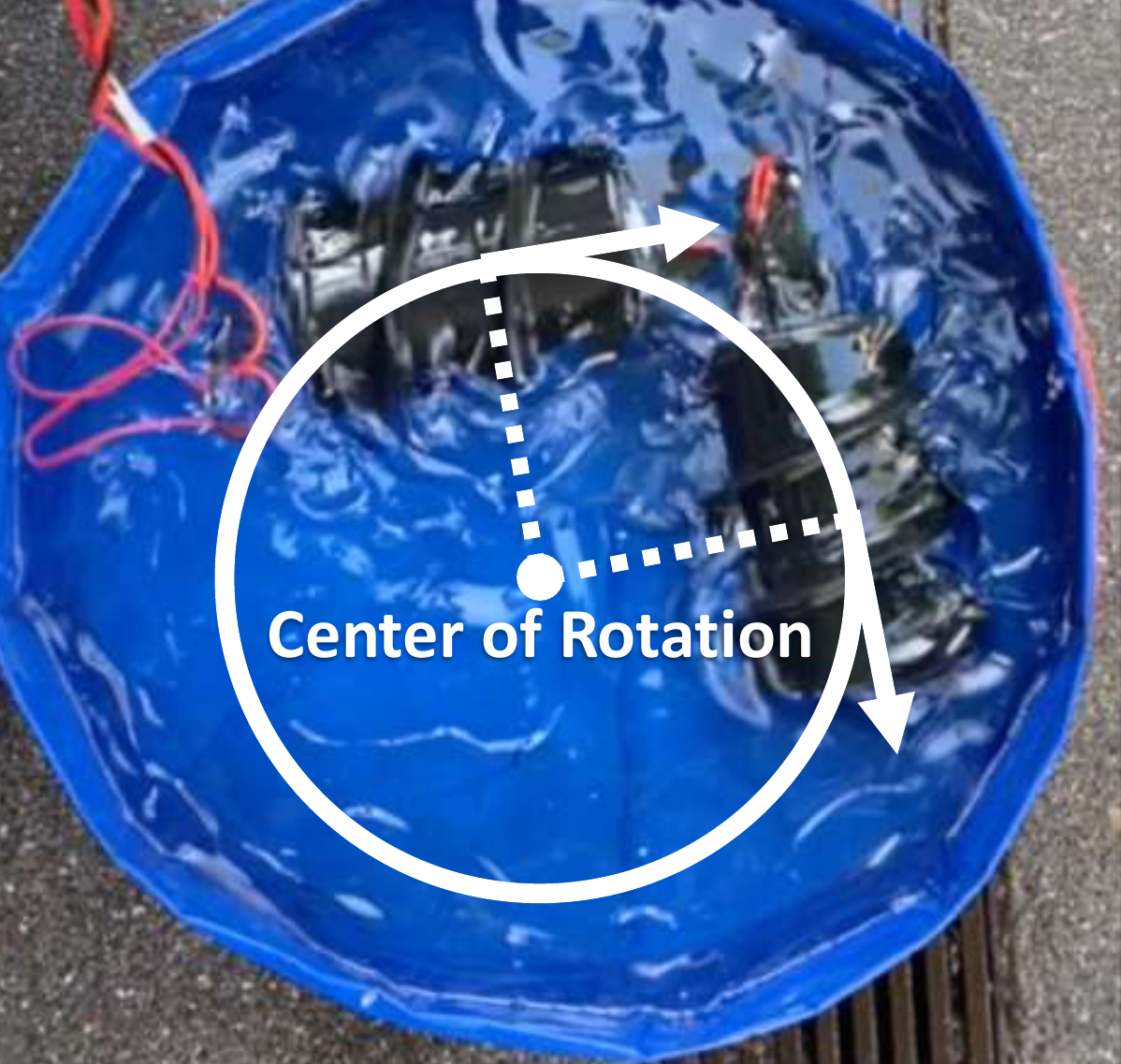}
    \includegraphics[trim=1.2cm 3.5cm 16.2cm 4cm, clip, width=0.527\columnwidth]{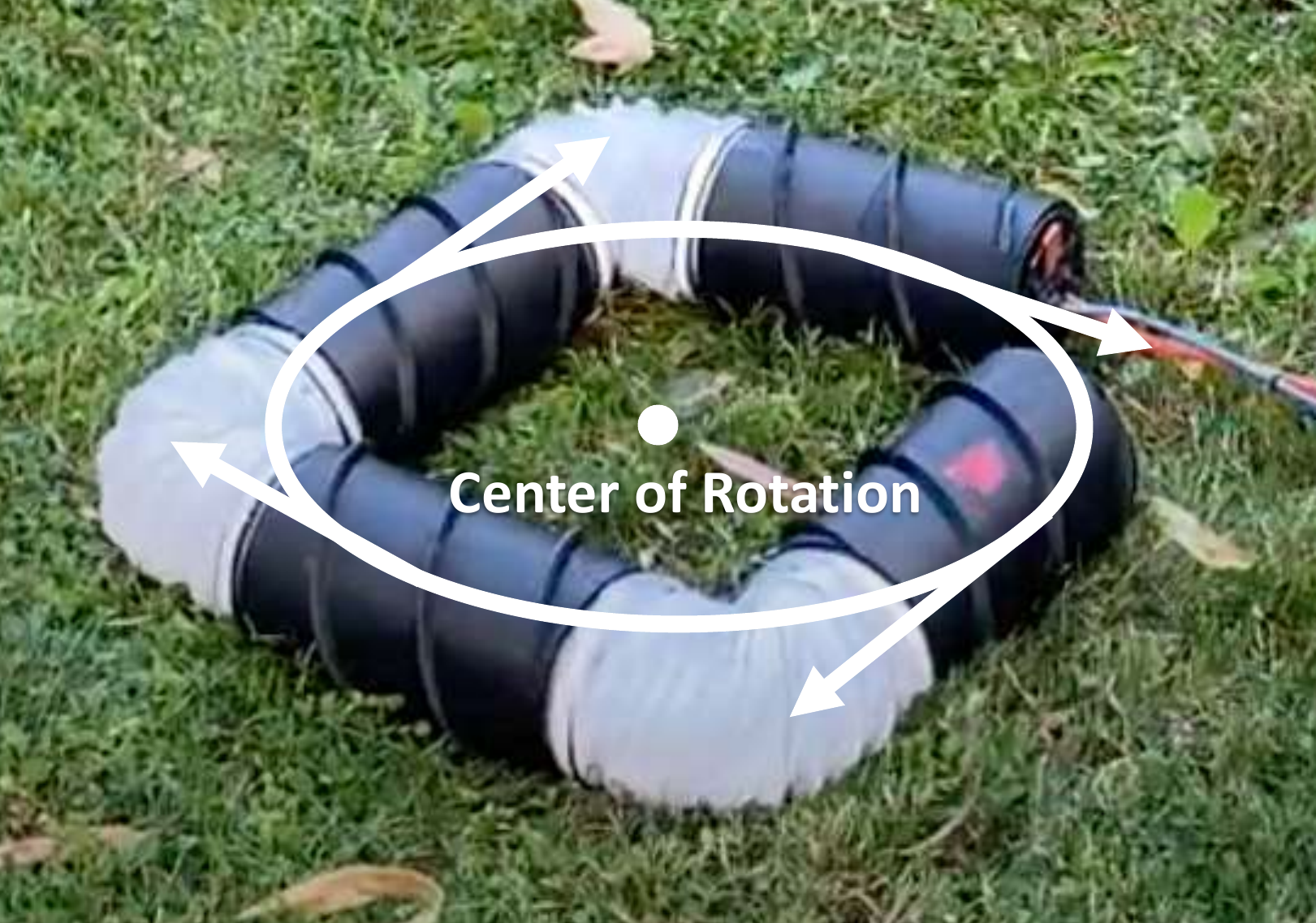}
    \includegraphics[trim=14.25cm 2cm 1cm 2.25cm, clip, width=0.94\columnwidth]{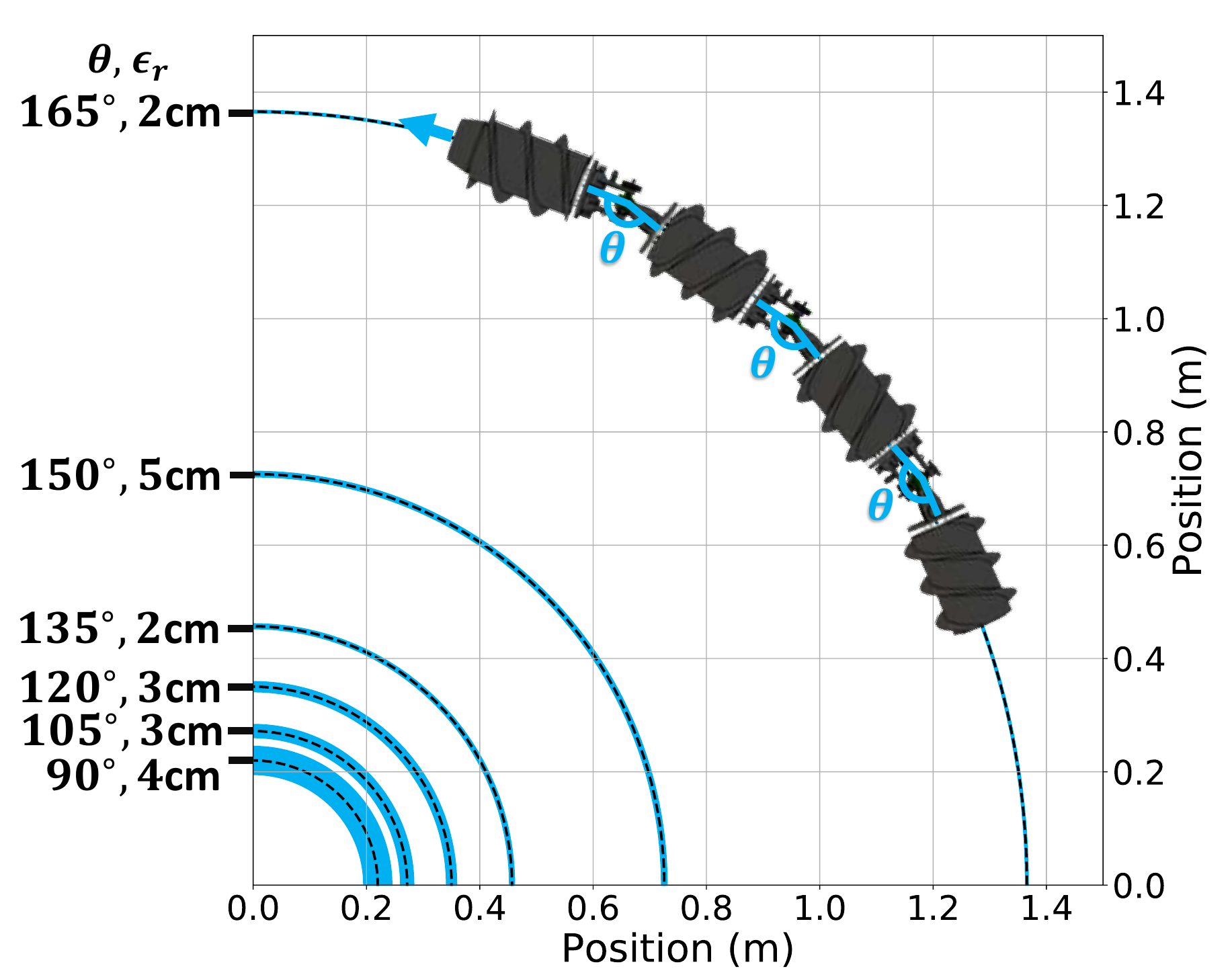}
    \caption{On screw-propellable media, the center of rotation can be controlled by adjusting the planar angle of all the U-Joints as shown in the top row of figures. This angle is denoted as $\theta$. ARCSnake’s U-Joints are set to various $\theta$ angles in tunneling mode on grass as shown in the top right figure. The mean, dashed line, and standard deviation, width of the blue line, of the measured turning radius are plotted here. The absolute difference between expected, calculated with the kinematic model, and mean measured turning radius, $\epsilon_r$, is also shown on the left-hand side of the plot.}
\label{fig:tunnelling_mode_kinematic_plot}
\end{figure}



    

Disturbances to the power line are of big concern since all the modules are daisy-chained.
To validate the power electronic design is robust to voltage disturbances, a single segment was placed in an inverted pendulum-like configuration.
The U-Joint was then regulated in a circular fashion, and the eventual positions were recorded at varying voltages.
A thermal image was also captured to ensure the heat losses from the power electronics is not of concern.
These results are shown in Fig. \ref{fig:regulating_different_voltages} which highlight the robustness of the electrical design.

The distribution of computations in the software architecture was also tested to ensure control loops can run effectively.
Each BeagleBone Black runs their main loop at 75Hz, which involves: reading IMU and encoder data, updating two PID controllers for the U-Joint, setting velocity for screw motor, broadcasting all sensor information, and receiving new set-points for the motors.
A round trip delay from sending a remote desktop command to the first daisy-chained BeagleBone Blacks and its returning sensor feedback was measured at $8.85\pm 0.11$ms over 1,000 samples.
This round trip delay was measured to increase on average by 0.31ms per BeagleBone along the daisy-chain.
These results mean that the software architecture can run a control loop with the remote desktop at 75Hz for up to 14 segments.
After 14 segments, there would be a scheduling concern for the data streams since the round-trip latency would be higher than the period ($\frac{1}{75}$s) of the control loop, so a lower rate may be required.
The coming experiments show that 75Hz is sufficient to produce effective control of ARCSnake.

\subsection{Tunneling Mode}

\begin{figure}[t]
    \centering
    \includegraphics[trim=2.6cm 0cm 2.6cm 0cm, clip, width=0.48\textwidth]{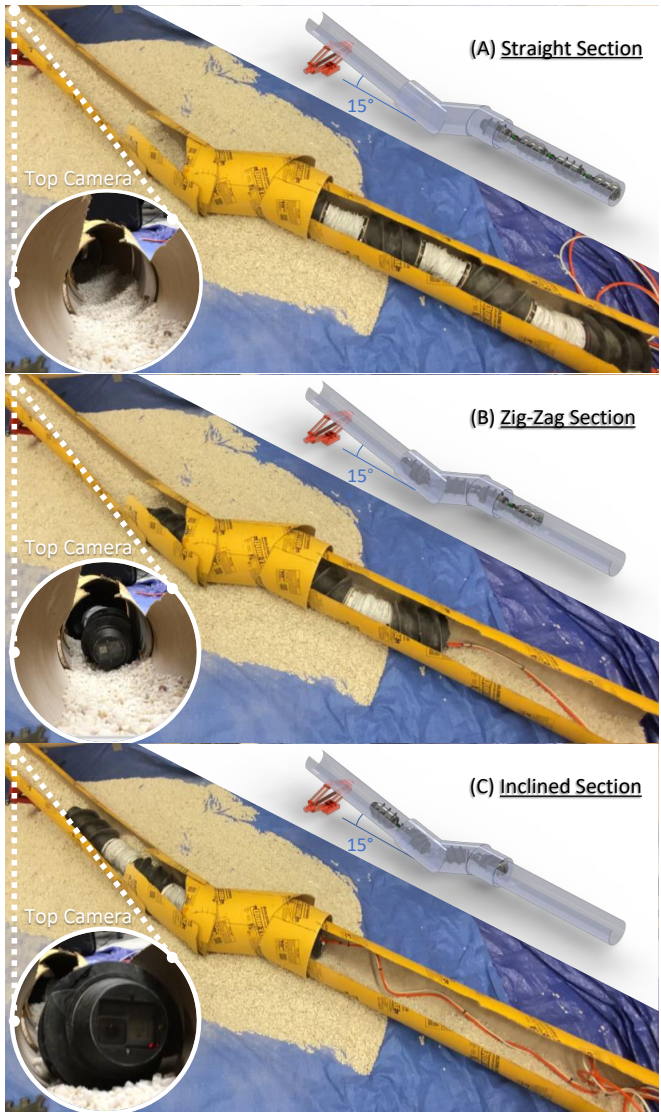}
    \caption{In a tortuous gravel course, ARCSnake tunnels through a narrow corridor that contains a (A) straight section, (B) zigzag section, and (C) 15$^\circ$ incline. Each section shows a model of the internal configuration (top-right) and a top camera view at the end of the inclined section (bottom-left). Only the head joint is actuated in this sequence to control the heading, while all other U-Joints are left in a back-drivable, complaint, mode. The back-drivable mode allows ARCSnake to naturally conform to the tortuous terrain.}
    \label{fig:cave_experiment}
\end{figure}

\begin{figure}[t]
    \centering
    \includegraphics[trim=0cm 5cm 16.25cm 4.1cm, clip, width=0.48\textwidth]{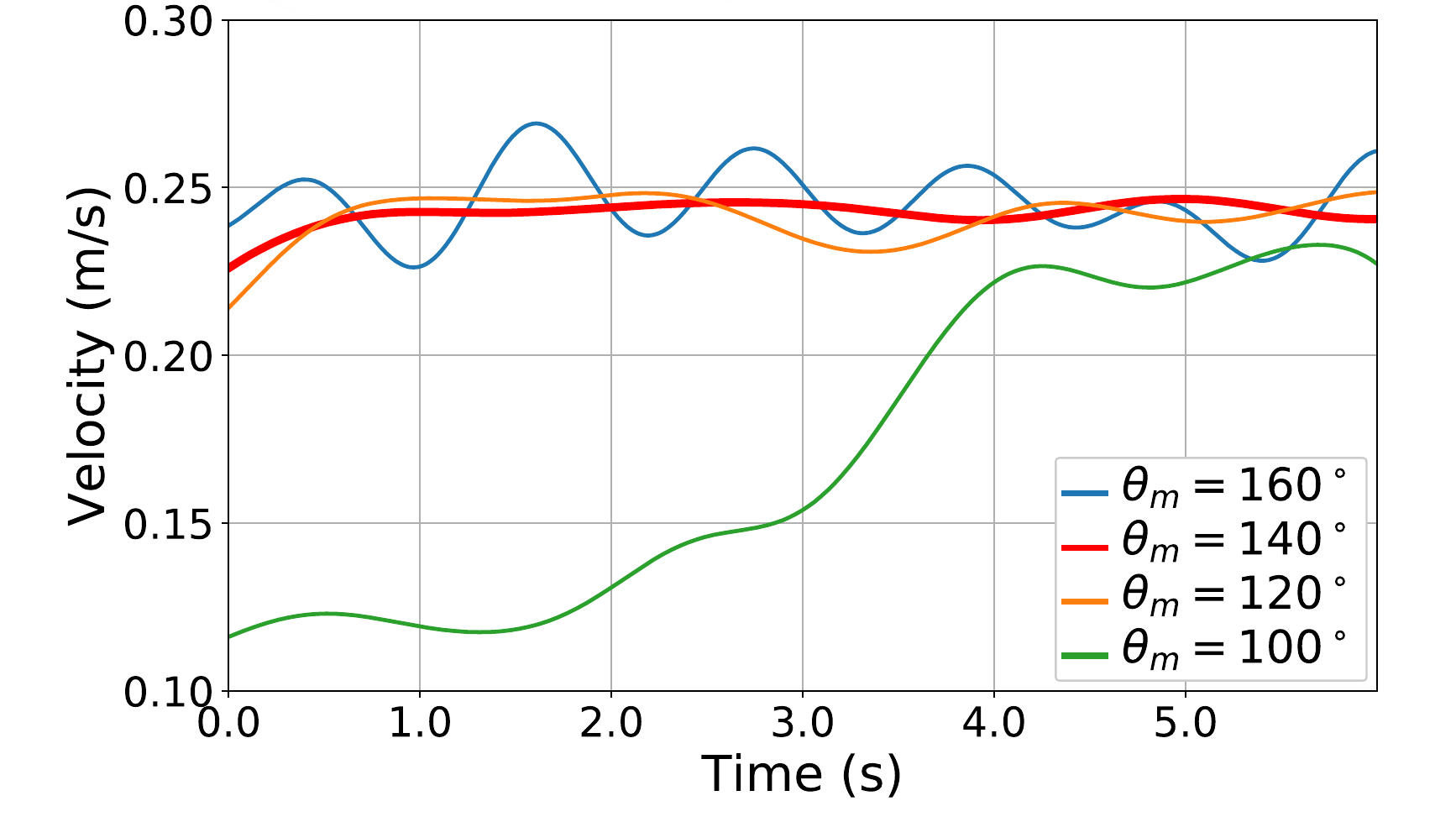}
    \caption{The instantaneous velocity is measured for various $\theta_m$ on tile flooring and the results are plotted here. 
    The M-configuration requires alternating planar U-Joint angles, $\theta_m$, to ensure there is enough leverage for propulsion.
    However, the locomotion becomes less efficient as the angle becomes steeper.
    In this experiment, the least oscillations are seen at $\theta_m=140^\circ$.}
    \label{fig:m_config_instantenous_vel}
\end{figure}

\begin{figure*}[t]
    \centering
    \includegraphics[trim=6.75cm 0cm 6.75cm 0cm, clip, width=0.24\textwidth]{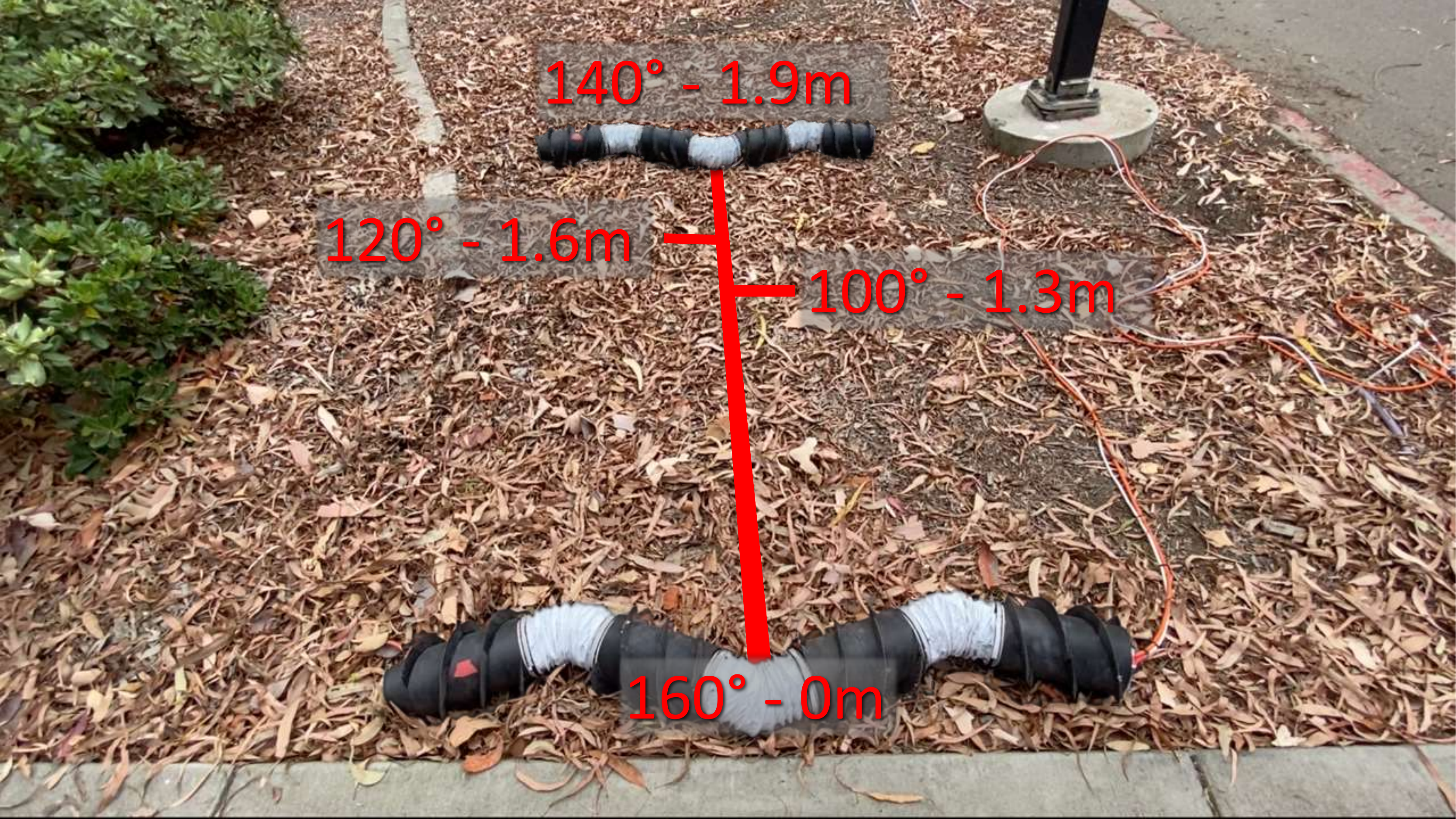}
    \includegraphics[trim=5.5cm 0cm 8.0cm 0cm, clip, width=0.24\textwidth]{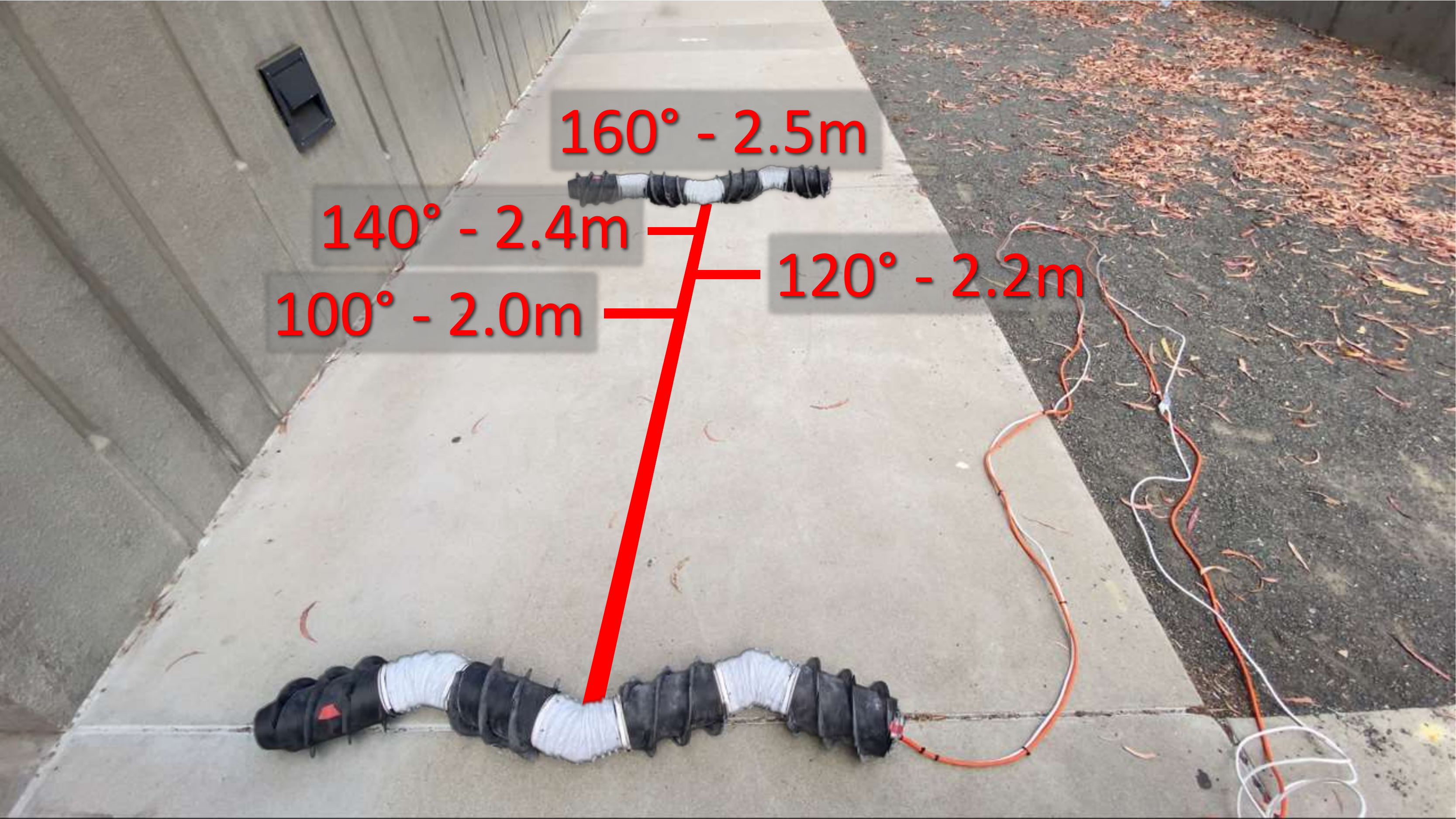}
    \includegraphics[trim=7cm 0cm 6.5cm 0cm, clip, width=0.24\textwidth]{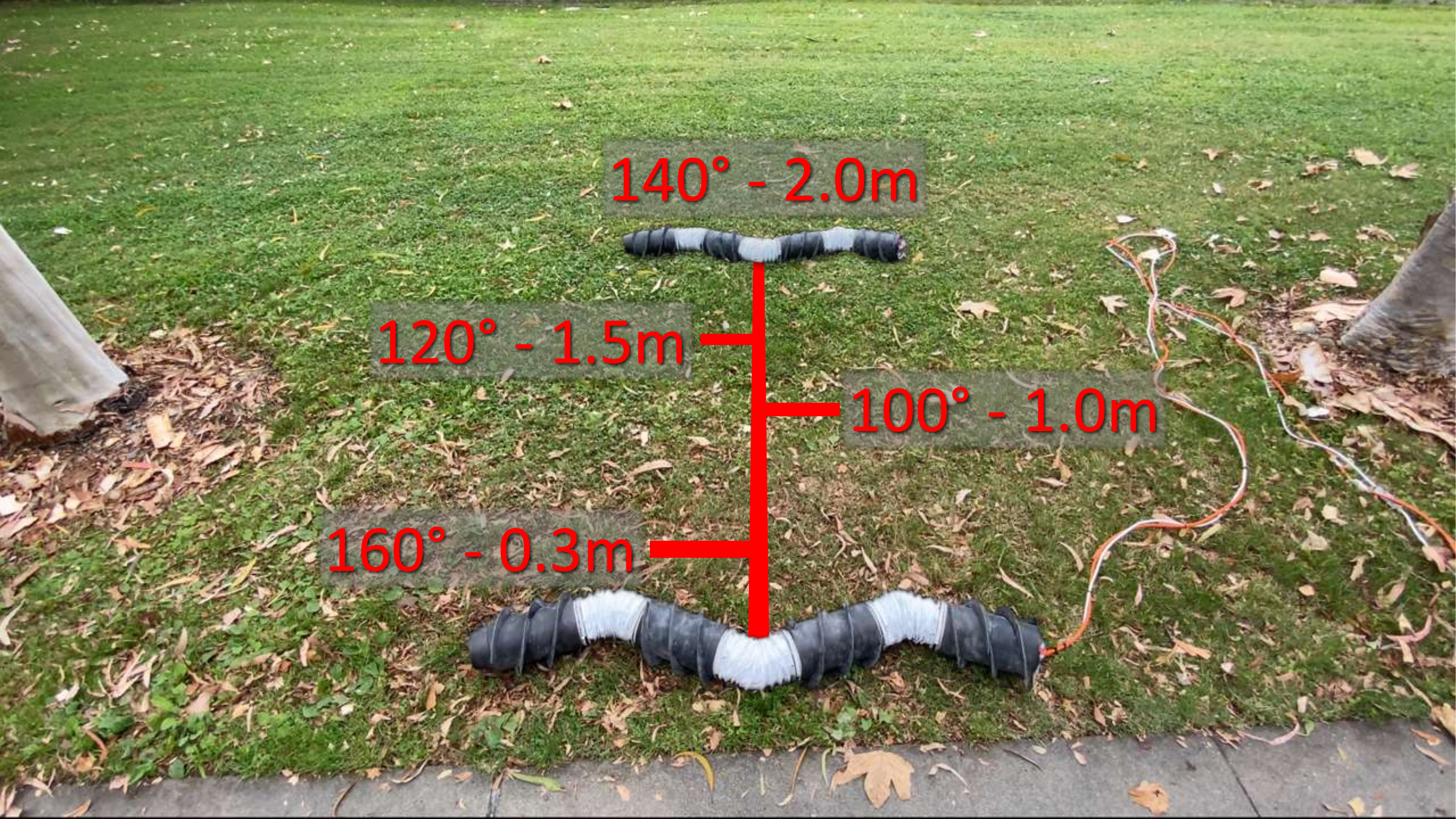}
    \includegraphics[trim=6.5cm 0cm 7cm 0cm, clip, width=0.24\textwidth]{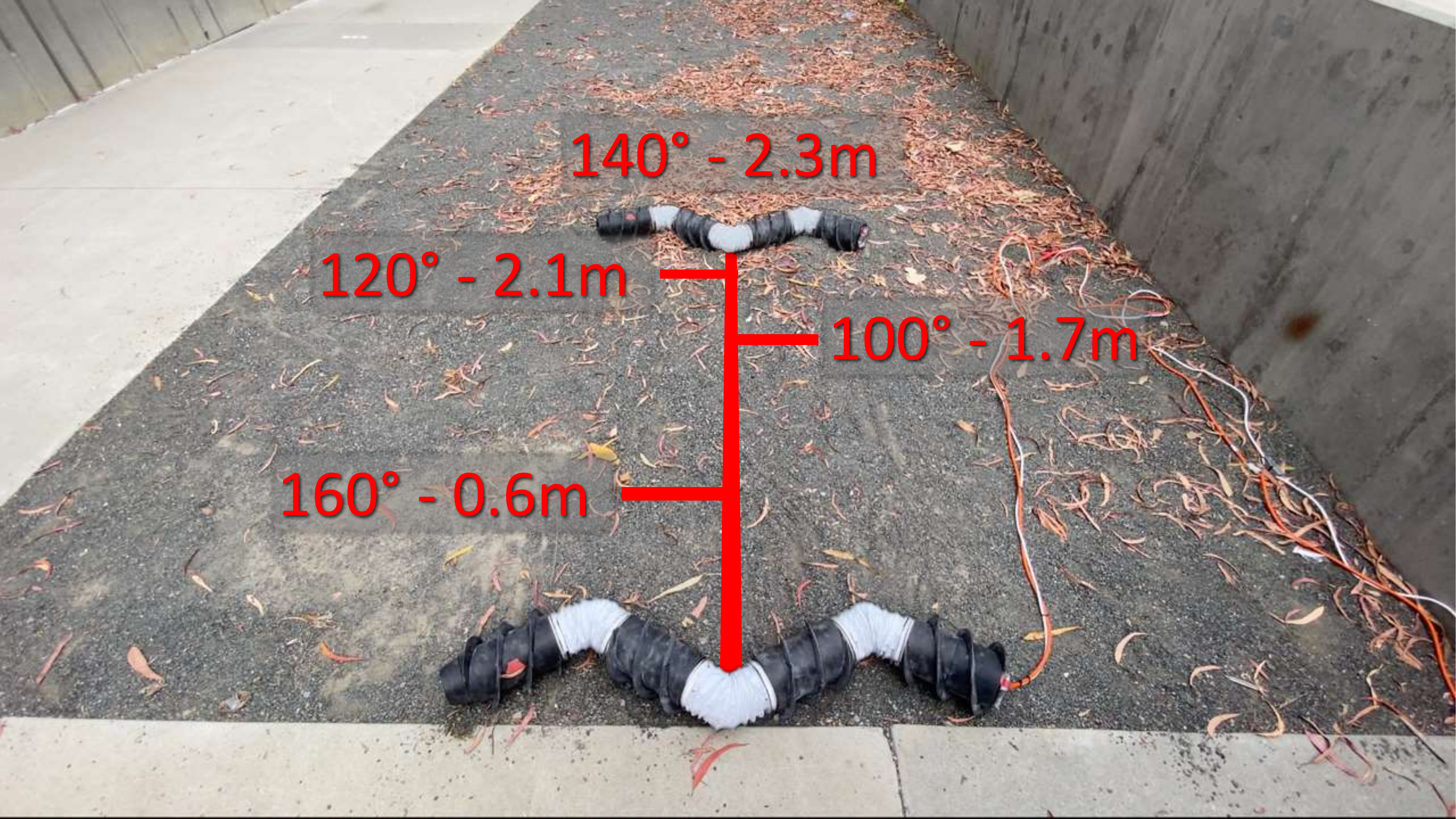}
    \caption{Total distance traveled in the M-configuration over 10 seconds with varying values of $\theta_m$ on four different surfaces. The surfaces from left to right are: forest floor, concrete, grass, and gravel. As seen by the results, $\theta_m = 140^\circ$ consistently travels the fastest.}
    \label{fig:m_config_outdoor_tests}
\end{figure*}

In open environments, ARCSnake is able to control its heading in tunneling mode by setting the center of rotation.
As shown in the kinematic modeling section, the center of rotation is the point where all the tangents of the screw segments' velocity intersect.
In order to ensure there is one intersection point, all planar U-Joint angles must be set to the same value, and only screw propulsion results on each segment.
The first condition is easy to meet as the U-Joints are actuated and can be regulated to target positions.
Meanwhile, the second condition is met when traversing on screw-propellable media which previously has been shown to be a wide range of terrain \cite{neumeyer1965marsh,dugoff1967model,fales1972riverine}.
This kinematic model was validated by setting the U-Joint angles and comparing the measured turning radius against its expected value based on the kinematic model.
The turning radius was measured using OptiTrack’s V120:Trio, and the experiment was conducted on outdoor grass.
The complete set of results are shown in Fig. \ref{fig:tunnelling_mode_kinematic_plot}, and the largest error measured occurred at a 43cm turning radius with an error of 5cm, 3.8\% of ARCSnake's total length.
Also, a minimum turning radius of 18cm, which is 14\% of ARCSnake's total length, was achieved in this mode.
This minimum turning radius was replicated in a small tub of mineral oil with a 2-link version of ARCSnake as also shown in Fig. \ref{fig:tunnelling_mode_kinematic_plot}.
This validation implies that ARCSnake has effective controllability of the heading in tunneling mode on any screw-propellable media when the U-Joints are not constrained by the environment.

For tortuous and narrow corridors, ARCSnake can utilize its hyper-redundancy to conform to the environment rather than control heading while tunneling.
However, directly controlling the U-Joints to conform is challenging because high-quality environmental mapping and localization would be required.
To circumvent this problem, the U-Joints are left in a back-drivable mode and conform to the environment through the Archimedean screw propulsion.
Only the head joint is actuated to guide the robot within the corridor.
As shown in Fig. \ref{fig:cave_experiment}, a cave experiment was conducted to test this.
ARCSnake, which has a cross-sectional diameter of 15cm including screw blades, successfully traversed through 20-25cm diameter corridors with tortuous turns and a 15$^\circ$ incline.

\subsection{M-Configuration}

The angle of the M-configuration, $\theta_m$, allows for a range of angles from 180$^\circ$ (straight) to 90$^\circ$.
The angle gives leverage for locomotion but causes the screws to fight one another hence being less efficient.
To understand this trade-off, the velocity when driving straight was measured using OptiTrack's V120:Trio across varying angles on indoor tile-flooring.
The results are shown in Fig. \ref{fig:m_config_instantenous_vel} with the screws running at 30\% of their maximum speed, and $\theta_m = 140^\circ$ oscillates the least.
The other angles measured oscillations from either limited leverage or the opposing screws fighting one another.
A similar experiment was run on other terrains in order to compare the performance across a variety of surfaces: concrete, flooring, gravel, grass, and forest floor.
For these tests, the distance travelled over 10 seconds was measured five times, and the mean results are shown in Fig. \ref{fig:m_config_outdoor_tests}.
The corresponding average velocities for this experiment are shown in Table \ref{tab:m_config_velocity_results}.
From the results, it is clear that the M-configuration angle, $\theta_m=140^\circ$ performs the most consistently with a max deviation of 0.05m/s across the measured surfaces.

\begin{table}[t]
    \centering
    \caption{The M-configuration’s speed measured for various U-Joint angles, $\theta_m$, on five different surfaces.}
    \def\arraystretch{1.2}
    \begin{tabular}{p{.2\linewidth}p{.125\linewidth}p{.125\linewidth}p{.125\linewidth}p{.125\linewidth}}
    \hline
      \textbf{Surface}  & $100^\circ$ & $120^\circ$ & $140^\circ$ & $160^\circ$\\ \hline
      Tile Flooring & 0.16 m/s & 0.22 m/s & 0.23 m/s & \textbf{0.24 m/s} \\
      Forest Floor & 0.13 m/s & 0.16 m/s & \textbf{0.19 m/s} & 0.00 m/s \\
      Concrete & 0.20 m/s & 0.22 m/s & 0.24 m/s & \textbf{0.25 m/s}\\
      Grass & 0.10 m/s & 0.15 m/s & \textbf{0.20 m/s} & 0.03 m/s \\
      Gravel & 0.17 m/s & 0.21 m/s & \textbf{0.23 m/s} & 0.06 m/s\\
        \hline
    \end{tabular}
    \label{tab:m_config_velocity_results}
\end{table}

\begin{figure}[t]
    \centering
    \includegraphics[trim=0cm 2.4cm 16.5cm 1.6cm, clip, width=0.46\textwidth]{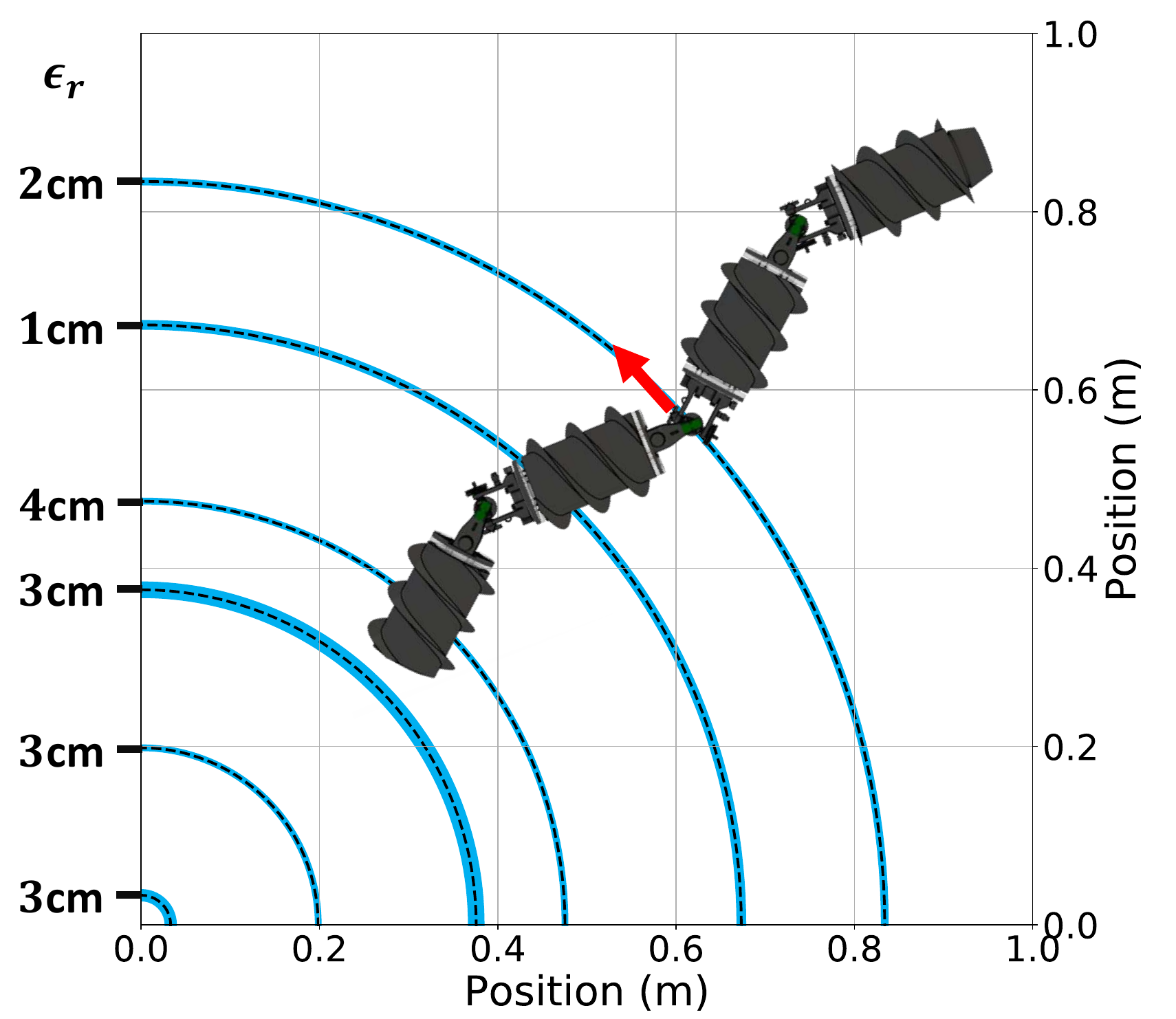}
    \caption{Various turning radii are targeted on a concrete surface in the M-configuration, and the mean, dashed line, and standard deviation, width of the blue line, of the measured turning radius are plotted here.
    The M-configuration controls the turning radius by adjusting screw speeds and the absolute error between the targeted and measured turning radius, $\epsilon_r$, is shown on the left-hand side.}
    \label{fig:m_config_turning_radius}
\end{figure}

To set the heading in the M-configuration, the speeds of the screws can be adjusted such that an instantaneous center of rotation is created.
To validate this kinematic model and highlight its robustness, a turning radius experiment was conducted on a concrete surface. 
The screws for ARCSnake are set to accomplish an expected turning radius and the actual turning radius was measured with OptiTrack's V120:Trio.
The results are shown in Fig. \ref{fig:m_config_turning_radius}.
The worst error in turning radius measured is 4cm at a targeted 51cm radius.
The minimum turning radius achieved is 0cm meaning ARCSnake can turn while staying in the same position.

\subsection{Snake-Inspired Maneuvers}

\begin{figure}
    \centering
    \includegraphics[width=0.48\textwidth]{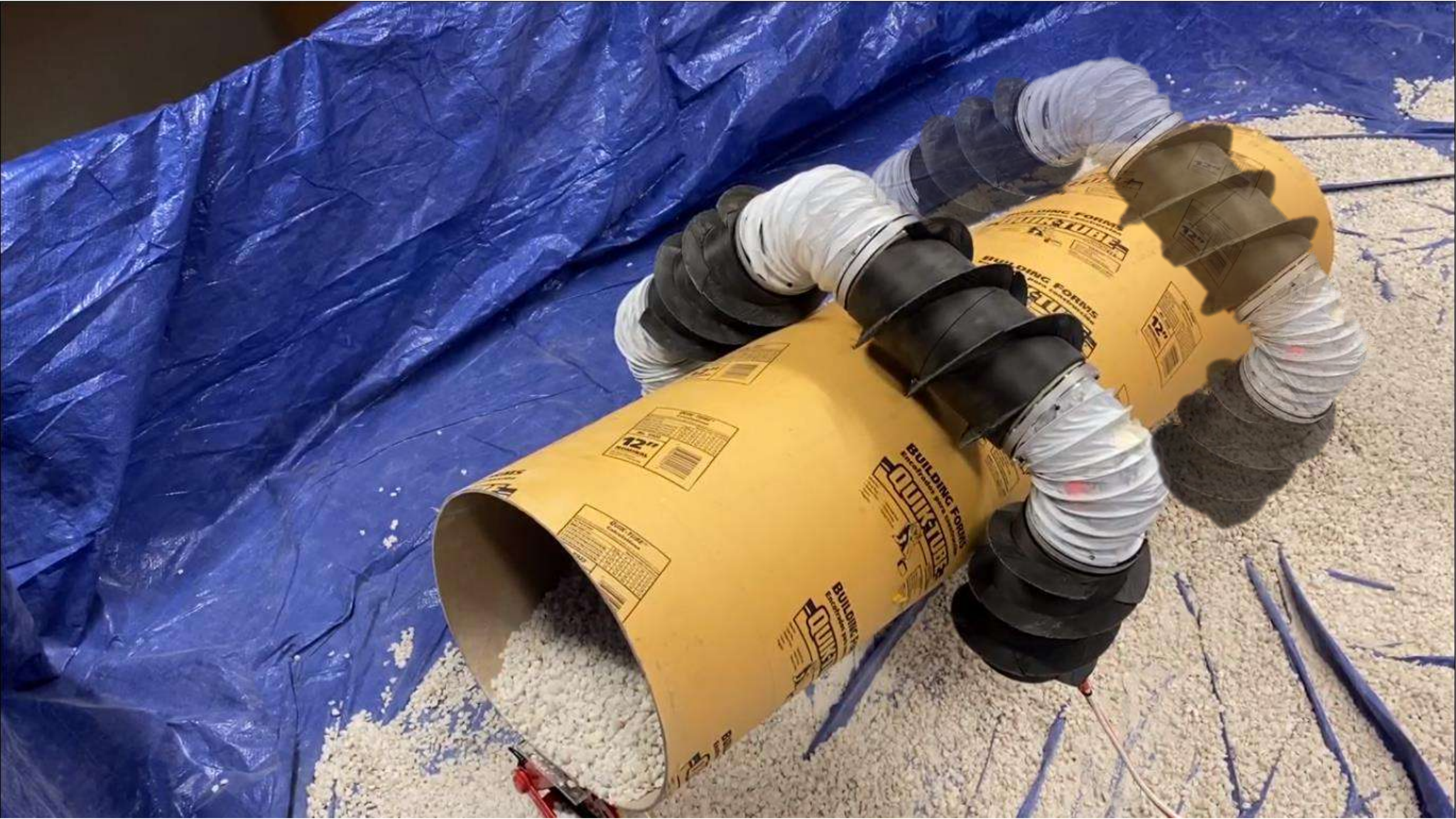}
    \caption{Similar to how biological snakes climb trees, ARCSnake can constrict itself onto an inclined pipe to maintain surface contact for grip and propulsion.}
    \label{fig:clamping}
\end{figure}

\begin{figure*}
    \centering
    \includegraphics[width=0.98\textwidth]{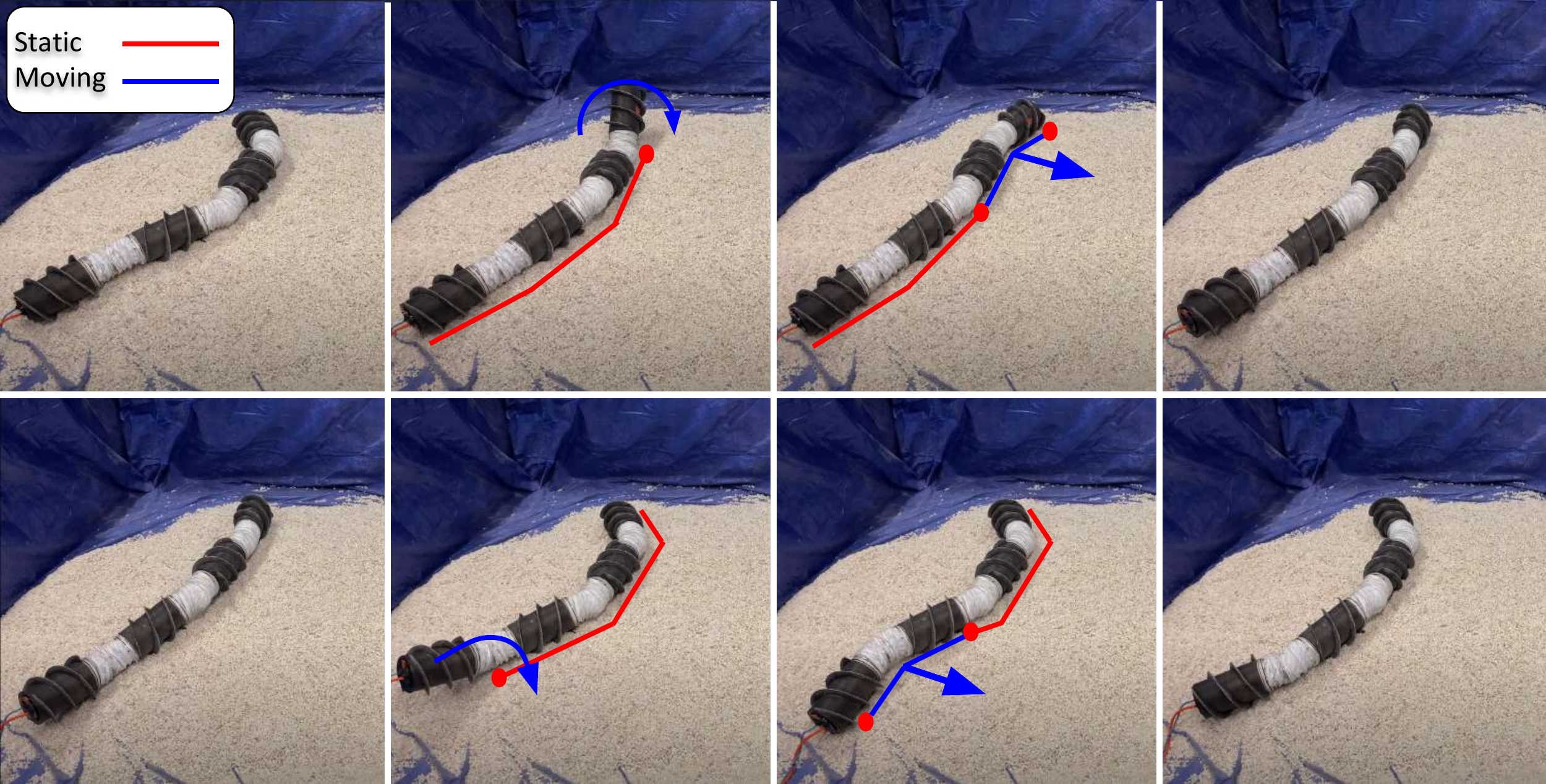}
    \caption{To achieve snake-like sidewinding, ARCSnake lifts and translates each segment in succession. Repeating this pattern allows ARCSnake to maneuver across loose surfaces.}
    \label{fig:side_winding}
\end{figure*}

\begin{figure*}
    \centering
    \includegraphics[trim=0cm 0cm 12.5cm 0cm, clip, width=0.32\textwidth]{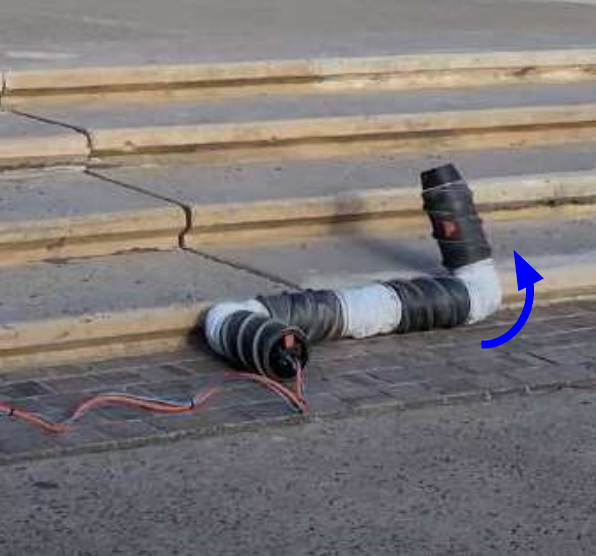}
    \includegraphics[trim=6.25cm 0cm 6.25cm 0cm, clip, width=0.32\textwidth]{Figures/snake_like/Climb_compressed.pdf}
    \includegraphics[trim=12.5cm 0cm 0cm 0cm, clip, width=0.32\textwidth]{Figures/snake_like/Climb_compressed.pdf}
    \caption{ARCSnake uses a combination of snake-inspired movements and Archimedean screw propulsion to climb up four 11.5 cm (92\% of body diameter) ledges in succession. }
    \label{fig:stair_climbing}
\end{figure*}

Operating in extreme and unknown environments includes unexpected complications in terrain, meaning a platform designed for them must come prepared with a wide library of locomotion strategies.
Taking advantage of ARCSnake’s hyper-redundancy, control strategies can leverage the biomimicry of biological snakes.
While not a comprehensive list of all possible snake-inspired control strategies possible on ARCSnake, a subset is conducted with Voodoo Doll to show the feasibility of additional maneuvers for locomotion.
Additionally, the screws can be utilized in tandem with these maneuvers to further enhance their capabilities. 

For the first example, some biological snakes climb tree trunks and branches by constricting themselves onto the surface to maintain grip \cite{arnold_2014}.
ARCSnake was able to mimic this ability by using its U-Joints to clamp on trunks as shown in Fig.  \ref{fig:clamping}.
In addition to bracing, ARCSnake was capable of locomoting along the trunk by using the screws for propulsion.
Note that screws had sufficient contact from the clamping forces supplied by the U-Joints.
This was tested on a 30cm diameter tube acting as a trunk at a 15$^\circ$ incline, and ARCSnake successfully maneuvers up and down the trunk. 

On terrains where there are flat surfaces lacking the projections needed to push off of, biological snakes sidewind by keeping portions of their bodies static while lifting and translating other portions of their body to a new location \cite{jayne1986kinematics}.
Similarly, ARCSnake can use 3D-joint movements to mimic sidewinding, as shown in Fig.  \ref{fig:side_winding}.
While in the case of climbing ledges, snakes will partition their body into three sections: on top of the ledge, free-hanging in the middle, and on the bottom of the ledge.
The top and bottom sections will use undulation to both anchor and propel the snake forward \cite{Gartjeb185991}.
ARCSnake replicated a similar maneuver, but instead of undulation, the screws were used to anchor and propel.
As shown in Fig. \ref{fig:stair_climbing}, ARCSnake successfully climbed four 11.5cm ledges in succession.
Note that each individual ledge is 92\% of ARCSnake’s body diameter.

\section{Discussion}

\begin{figure}[t]
    \centering
    \vspace{2mm}
    \includegraphics[trim=0.8cm 2.5cm 2cm 2cm, clip, width=0.48\textwidth]{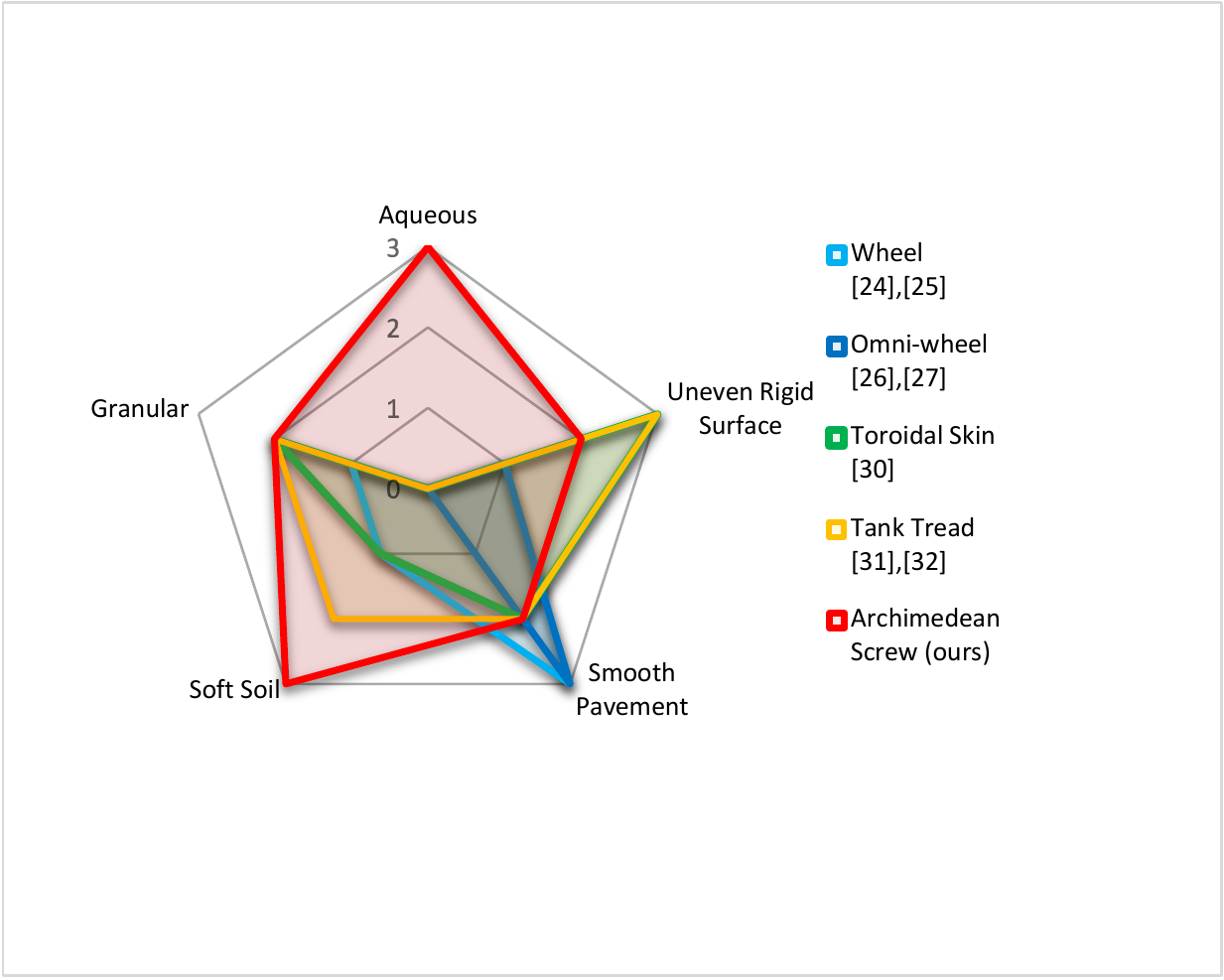}
    \caption{The plot shows a comparison of the locomotion effectiveness of previously developed active skins and our proposed Archimedean Screw in various domains. Screw-propellable media, such as soft soil, aqueous, and granular surfaces, are where previous active skins fail to effectively locomote. Our proposed active skin, the Archimedean screw, is able to locomote on these challenging environments as shown in our tunneling mode experiments. On more rigid surfaces, such as pavement, the Archimedean screw is still able to provide effective locomotion as shown in the M-configuration experiments. Therefore, the Archimedean screw as an active skin for snake robots enables multi-domain mobility.}
    \label{fig:active_skin_effectiveness}
\end{figure}


The robustness of the mechanical, electrical, and software design for ARCSnake enabled the high number of experiments conducted on the platform.
The first set of mobility experiments showed that by utilizing the Archimedean screws, ARCSnake can traverse through screw-propellable media and requires only space for its cross-section.
The screw-propellable media tested on were: gravel, mineral oil, and grass.
A kinematic relationship between the U-Joint angles and heading was derived and validated, hence allowing for effective control of ARCSnake in this mode.
Furthermore, in cases where ARCSnake is highly constrained due to the environment such as pipes and tunnels, the U-Joints can be left in a back-drivable mode and will naturally conform to the environment.
When the screws are slipping, the Archimedean screws will act more similar to that of a wheel, and ARCSnake can reconfigure to the M-configuration to still have consistent locomotion.
To ensure the robustness of the M-configuration, it was tested on tile flooring, forest floor, concrete, grass, and gravel; all of which result in screw slippage.
Similar to tunneling mode, a kinematic relationship between screw speed and heading was derived and validated.

A comparison of the effectiveness of previously developed active skins for snake robots and ours is shown in Fig. \ref{fig:active_skin_effectiveness}.
The different active skins performance are categorized according to the following locomotion capabilities on the corresponding domain: 
\begin{itemize}
    \item $0$ implies it cannot provide mobility
    \item $1$ implies it can provide mobility
    \item $2$ implies it is designed to provide mobility
    \item $3$ implies it excels at providing mobility
\end{itemize}
The comparison highlights the multi-domain capability of ARCSnake.
This is enabled by the Archimedean screw and our novel configurations for it: tunneling mode for screw propellable media and M-configuration under screw-slippage.

The hyper-redundancy from the serially chained U-Joints also allows ARCSnake to use control strategies similar to that of biological snakes.
By using Voodoo Doll, we demonstrated three out of the many potential control techniques possible: trunk clamping, side-winding, and ledge climbing.
It is not feasible to experiment for every possible terrain or obstacle potentially faced in exploration and search and rescue missions.
Nonetheless, the subset of surfaces tested provides a good spread across density, granularity, and fluidity, whilst the terrains tested involve challenges in elevation, tortuosity, and shape.
Therefore, the experiments are a representative subset of challenging environments.
The success of these tests showcases ARCSnake’s multi-domain mobility, providing confidence that ARCSnake can also challenge other types of environments.

ARCSnake also served as the first proof-of-concept demonstration for NASA's Exobiology Extant Life Surveyor (EELS) program.
The goal of the EELS program is to deliver scientific instrumentation deep within the plume vents, caves, and ice sheets of Enceladus and Europa in search of extant life.
We anticipate that Archimedean screw propulsion would be an effective form of locomotion when navigating through the tortuous caves.
Furthermore, the snake-like, hyper-redundancy can be used to descend down the plume vents in a similar fashion to the snake-inspired ledge experiment.
The back segments would anchor the robot while the head segments will descend down the plume vents.
Finally, the wide breadth of locomotion strategies possible with ARCSnake will be beneficial for the many unforeseen challenges in the mission.

\section{Conclusion}
Exploration missions in complex environments inherently require a highly adaptive platform due to the unknown terrain and obstacles which are too treacherous or inaccessible for human access.
The wide breadth of scenarios that ARCSnake can handle makes it an ideal candidate for these extreme environments.
ARCSnake can traverse narrow, tortuous channels through Archimedean screw propulsion.
In cases of rigid, hard surfaces where screw propulsion fails, ARCSnake can utilize its kinematic hyper-redundancy to still produce effective locomotion.

For future work, we intend to better model the interaction between Archimedean screws and surfaces.
Previous work has modeled isolated screw propulsion effects \cite{nagaoka2010terramechanics}, and we would utilize this as a starting point for our model.
The modeling will also incorporate U-Joints which can be maneuvered to adjust surface contact.
The end result would lead to potentially more efficient forms of screw-propelled locomotion.

\section*{Acknowledgements}
This work was funded by the NASA Jet Propulsion Laboratory under the 2018-2019 Spontaneous Concept Award. 
F. Richter is supported via the National Science Foundation Graduate Research Fellowships.
The authors would like to thank Kalind Carpenter for his conceptual and technical support and Casey Price for her help in software development on the ARCSnake platform.

\bibliographystyle{ieeetr} 
\bibliography{references}

\end{document}